\documentclass[preprint,12pt,authoryear]{elsarticle}

\usepackage{amssymb}

\usepackage{float}
\usepackage{tikz}
\usepackage{subcaption}
\usepackage{caption}
\usepackage{amsmath}
\usepackage{booktabs}
\usepackage{multirow}

\begin{document}

\begin{frontmatter}



\title{Forest aboveground biomass estimation using GEDI and earth observation data through attention-based deep learning}


\author[inst1]{Wenquan Dong}
\author[inst1,inst2]{Edward T.A. Mitchard}
\author[inst3]{Hao Yu*}
\author[inst1]{Steven Hancock}
\author[inst1]{Casey M. Ryan}

\affiliation[inst1]{organization={School of GeoSciences, University of Edinburgh},
            addressline={King's Buildings}, 
            city={Edinburgh },
            postcode={EH9 3FF}, 
            country={UK}}
\affiliation[inst2]{organization={Space Intelligence Ltd},
            addressline={93 George Street}, 
            city={Edinburgh },
            postcode={EH2 3ES}, 
            country={UK}}       
            
\affiliation[inst3]{organization={School of Engineering, University of Edinburgh},
            addressline={King's Buildings}, 
            city={Edinburgh },
            postcode={EH9 3FG}, 
            country={UK}}
            
\begin{abstract}
Accurate quantification of forest aboveground biomass (AGB) is critical for understanding carbon accounting in the context of climate change. In this study, we presented a novel attention-based deep learning approach for forest AGB estimation, primarily utilizing openly accessible Earth Observation (EO) data, including: Global Ecosystem Dynamics Investigation (GEDI) LiDAR data, C-band Sentinel-1 synthetic-aperture radar (SAR) data, Advanced Land Observing Satellite-2 (ALOS-2) Phased Array type L-band Synthetic Aperture Radar-2 (PALSAR-2) data, and Sentinel-2 multispectral data. The attention UNet (AU) model achieved markedly higher accuracy for biomass estimation compared to the conventional random forest (RF) algorithm. Specifically, the AU model attained an R$^2$ of 0.66, RMSE of 43.66 Mg ha\textsuperscript{-1}, and bias of 0.14 Mg ha\textsuperscript{-1}, while RF resulted in lower scores of R$^2$ 0.62, RMSE 45.87 Mg ha\textsuperscript{-1}, and bias 1.09 Mg ha\textsuperscript{-1}. However, the superiority of the deep learning approach was not uniformly observed across all tested models. ResNet101 only achieved an R$^2$ of 0.50, an RMSE of 52.93 Mg ha\textsuperscript{-1}, and a bias of 0.99 Mg ha\textsuperscript{-1}, while the UNet reported an R$^2$ of 0.65, an RMSE of 44.28 Mg ha\textsuperscript{-1}, and a substantial bias of 1.84 Mg ha\textsuperscript{-1}. Moreover, to explore the performance of AU in the absence of spatial information, fully connected (FC) layers were employed to eliminate spatial information from the remote sensing data. AU-FC achieved intermediate R$^2$ of 0.64, RMSE of 44.92 Mg ha\textsuperscript{-1}, and bias of -0.56 Mg ha\textsuperscript{-1}, outperforming RF but underperforming AU model using spatial information. We also generated 10m forest AGB maps across Guangdong for the year 2019 using AU and compared it with that produced by RF. The AGB distributions from both models showed strong agreement with similar mean values; the mean forest AGB estimated by AU was 102.18 Mg ha\textsuperscript{-1} while that of RF was 104.84 Mg ha\textsuperscript{-1}. Additionally, it was observed that the AGB map generated by AU provided superior spatial information. Overall, this research substantiates the feasibility of employing deep learning for biomass estimation based on satellite data.

\end{abstract}

\begin{highlights}

\item Deep learning methods for forest AGB mapping using GEDI and earth observations.
\item Attention UNet (AU) outperforms conventional random forest model.
\item The AU generated AGB maps with superior spatial detail compared to the random forest model.
\item AU's enhanced accuracy is due to effective utilization of spatial information and its inherent algorithmic mechanisms.
\item 10m AGB map for Guangdong, China in 2019.

\end{highlights}


\end{frontmatter}


\section{Introduction}
Forest ecosystems are fundamental to the global carbon cycle, holding 80\% of the Earth's total vegetation biomass \citep{kindermann2008global}. Notably, of this forest biomass, 80\% is stored above ground \citep{pan2013structure}. Forest aboveground Biomass (AGB) have garnered significant attention given their profound implications for environmental and climate science \citep{mitchard2018tropical}. The accurate quantification and monitoring the dynamics of forest AGB are instrumental in enhancing our understanding of the global carbon cycle. Such insights provide a robust foundation for the formulation of evidence-based policies geared towards sustainable forest management and climate change mitigation \citep{herold2019role}.

Estimation methodologies for forest biomass are predominantly classified into two main categories: in-situ field measurements and remote sensing techniques. The most accurate, yet time-consuming small-scale approach for assessing AGB is based on in-situ destructive sampling \citep{kumar2017remote}. Alternatively, field measurements typically involve selecting representative sites and measuring tree canopy height and diameter at breast height (DBH) using specialized instruments \citep{mitchard2009using}. In certain studies, wood density are also measured \citep{chave2005tree}. Subsequently, these empirical measurements are integrated into allometric equations to derive biomass estimates \citep{chave2014improved}. It's noteworthy, however, that such field-based methodologies, while precise, are labor-intensive and inherently limited in spatial coverage, often constrained to areas that are logistically feasible to access. In contrast, remote sensing techniques offer the advantages of large-scale monitoring combined with high spatiotemporal resolution \citep{saatchi2011benchmark, santoro2021global}. Spaceborne remote sensing serves as an important approach for large-scale monitoring of AGB, providing explicit spatial detail and coverage, enabling high-resolution surveys of relatively extensive areas at a lower cost.\citep{baccini2012estimated, brandt2018satellite}. Optical and radar data serve as primary sources for estimating forest AGB. Optical data provides rich spectral information about vegetation properties based on reflection, while SAR penetrates into the canopy to derive canopy structure parameters \citep{mitchard2011measuring, liu2019estimation, gomez2014historical}. However, they do not provide direct measurements of biomass \citep{woodhouse2012radar, campbell2021scaled}. Conversely, LiDAR technology offers the capability for direct measurement of tree heights, which are closely related to forest AGB. However, spaceborne LiDAR data is discontinuous, and as such, it is often integrated with optical and SAR data for forest AGB estimation.

Conventional machine learning techniques, such as Random Forest (RF),  Maximum Entropy Modeling (MaxEnt), and Support Vector Machines (SVM), have been widely used for extrapolating discontinuous LiDAR data to generate wall-to-wall maps through integration with optical and SAR data \citep{saatchi2011benchmark, liang2023quantifying, dhanda2017optimizing}.While these conventional machine learning models have shown promising results on forest AGB estimation using remote sensing data, they still have some inherent limitations that constrain their performance. Specifically, conventional machine learning techniques for remote sensing analysis often rely solely on spectral features extracted from individual samples \citep{reichstein2019deep}. Each pixel is treated as an independent data instance without considering inter-pixel spatial relationships within the imagery. However, forest AGB accumulation normally has an intrinsic spatial dependence that cannot be captured by treating pixels and samples independently. Neighboring pixels of forest exhibit spatial autocorrelation in AGB due to ecological factors and biophysical processes with spatial domains \citep{ploton2020spatial}. For instance, environmental variables like soil fertility and water availability that affect forest growth often have spatial patterns tied to topography and precipitation \citep{lewis2013above}. Therefore, accurate estimation of forest AGB requires characterization of these complex spatial patterns and relationships between neighboring pixels, rather than simplifying pixels as spatially independent data points. This suggests there is ample room for improvement by incorporating deep learning techniques, such as convolutional neural networks, graph neural networks and transformer models, which can automatically learn spatial features and patterns without relying on handcrafted features \citep{gu2018recent, zhou2020graph, dosovitskiy2020image}.

In the past few years, the field of remote sensing has begun to use deep learning to better exploit spatial and temporal information within datasets, achieving remarkable results in tasks such as classification and segmentation \citep{brandt2020unexpectedly, tong2023enabling, mpakairi2023fine}. The performance of deep learning largely depends on the availability and diversity of labeled datasets. An inadequate volume of data can result in model overfitting, while a lack of data diversity might compromise the model's generalization capacity \citep{long2021creating}. Unfortunately, expansive and labeled datasets are not always readily available. Typically, for classification and detection tasks, manual annotations are made through visual interpretation \citep{brown2022dynamic}. This approach, however, is not suitable for tasks like biomass estimation, as it is challenging to directly annotate the amount of biomass from remote sensing imagery. The advent of GEDI has opened up possibilities for employing deep learning in biomass estimation. The GEDI instrument conducts high-resolution lidar observations of the Earth's three-dimensional structure, enabling precise inversion of canopy height \citep{liu2021performance}. The GEDI instrument integrates three lasers, with one split into two "coverage" beams and the others emitting full "power" beams. At any instance, four beams with a 25 m footprint diameter measure the ground, and these are alternately dithered across tracks, resulting in eight data tracks within a 4.2 km swath. Each data track has footprint centers distanced at 60 m along the track \citep{dubayah2020global}. With its unprecedented dense sampling as a lidar instrument in orbit, GEDI can serve as an ideal labeled dataset to facilitate deep learning for forest attribute estimation.

In this study, we developed a novel approach to estimate forest AGB accurately through the application of deep learning techniques to remote sensing data. We established a connection between field data and GEDI RH, thereby converting GEDI data into AGB to serve as ground-truth labels. The framework we proposed relies on a UNet architecture incorporated with an attention mechanism, serving as the baseline model. Guangdong province, China, was selected as a case study region to demonstrate the applicability of the proposed models across complex forest landscapes. The specific objectives of the study were as follows: \romannumeral 1) to explore and validate the capabilities of deep learning methods for forest AGB estimation through the integration of multi-source remote sensing data; \romannumeral 2) to compare the results derived from deep learning methods with those obtained using the conventional RF method; \romannumeral 3) to analyze whether the improvements of AU over the RF method is due to AU’s utilization of spatial information.

\section{Data and methods}

\subsection{Study area}
Guangdong province, situated in the southernmost region of mainland China, encompasses a total area of approximately 179,725 square kilometers. It boasts a diverse topography, ranging from coastal plains in the south to mountainous regions in the north, and is bordered by the South China Sea to its south. The province's climate is predominantly subtropical, characterized by mild winters and hot, humid summers \citep{cheng2021extreme}. In Guangdong province, climatic conditions exhibit a seasonal variability. Mean monthly temperatures range from approximately 16-19°C in January to 28-29°C in July \citep{tian2022dynamic}. The province is subject to substantial precipitation, with average annual cumulative rainfall varying between 1,300-2,500 mm across regional gradients, fostering rich biodiversity and a variety of forest ecosystems. Guangdong is home to several forest types, including subtropical evergreen broad-leaved forests and monsoon rainforests. The area has undergone significant anthropogenic changes over the decades, marked by rapid urbanization and shifts in land use. From 1980 to 2020, the forested area in Guangdong province has experienced a noteworthy expansion, growing from 59,840 km\textsuperscript{2} to 105,241 km\textsuperscript{2} \citep{tian2022dynamic}. Meanwhile, the province's forestry industry maintained its leading position in the nation in terms of total output value for several consecutive years \citep{zhang2018ecological}. Thus, Guangdong province has been selected as the case study for this research, aiming to delineate a high-resolution forest AGB map using remote sensing data to aid sustainable forest management and conservation efforts.

\begin{figure}[H]
    \centering
    \begin{subfigure}{1.0\textwidth}
        \centering
        \begin{tikzpicture}
            \node[anchor=south west,inner sep=0] (image) at (0,0) {\includegraphics[width=\textwidth]{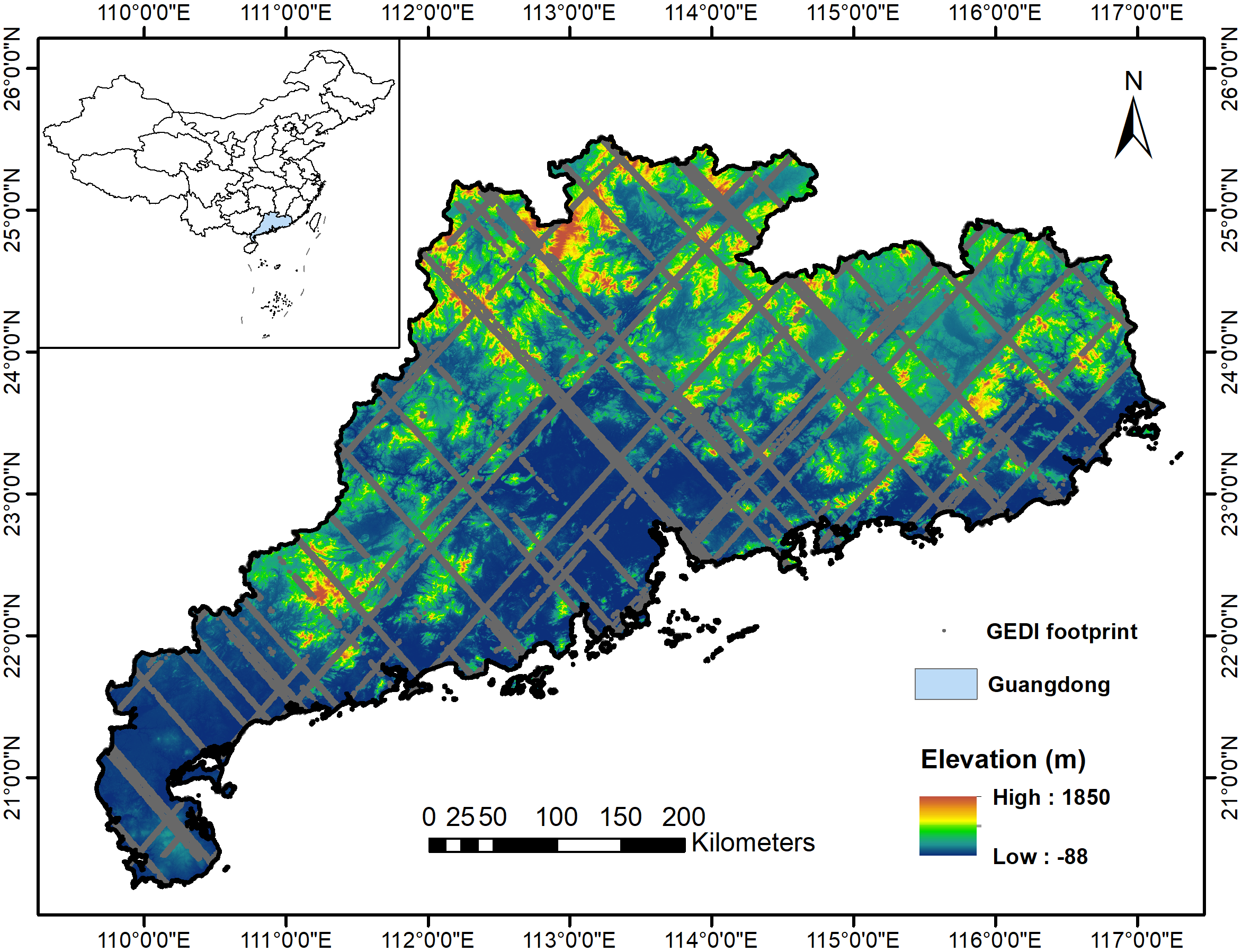}};
        \end{tikzpicture}
    \end{subfigure}%
\caption{The location of Guangdong province and the spatial distribution of GEDI footprints shown over a Digital Elevation Model (DEM), the NASADEM \citep{jpl2020nasadem}. }
\label{fig: Study area}
\end{figure}

\subsection{Data}

\subsubsection{GEDI}
GEDI Level 2A (L2A) and Level 2B (L2B) Version 2 data collected in 2019 were used in this study. GEDI L2A data provide footprint-level relative height (RH) metrics and ground elevation derived from Level 1 waveform data. \citep{dubayah2021gedi}. RH metric refers to the height at which a specified quantile of the cumulative energy of the return signal is achieved. By capturing the full information of height distribution, the RH metrics quantify the vertical profile of forest within a GEDI footprint, and thus strongly correlate with AGB density and provide key inputs into AGB estimation models using GEDI data \citep{dubayah2020global}. GEDI L2B data includes canopy cover for each footprint. Canopy cover is a key metric in forest ecology, refers to the percentage of the ground area that is obscured by the vertical projection of the tree canopy.

The GEDI L2A and L2B data were accessed via Google Earth Engine (GEE) \citep{gorelick2017google}. Initially, we queried the GEDI L2 table index specific to our study area. Subsequent to this query, we utilized the retrieved GEDI file names to obtain the GEDI L2 vector data. Both the GEDI L2A and L2B data provide parameters that can be used to filter GEDI footprints with low quality. GEDI footprints that were acquired during the day (solar elevation $\geq$ 0), or those that were degraded (degrade $=$ 1) or of low quality (quality flag $=$ 0), were deemed unreliable and excluded from the analysis. For the GEDI footprints with GEDI L2B canopy cover of less than 0.8, the minimal sensitivity used in this study is 0.9. Conversely, for the footprints with a canopy cover of 0.8 or greater, the minimal sensitivity is 0.98. In this study, only power beams were used because their use results in higher accuracy when estimating tree height and biomass \citep{liu2021performance, duncanson2020biomass}. The Version 2 GEDI data, compared to Version 1, boast significant improvements in geolocation accuracy, with a mean geolocation uncertainty of 10.3 m. However, an uncertainty of 10.3 m is still substantial for our AGB map with a 10 m resolution. To reduce the geolocation uncertainty, footprints were filtered based on two criteria: firstly, where the absolute difference between the GEDI L2B canopy cover and the kNDVI derived from Landsat exceeded one standard deviation from the mean \citep{liang2023quantifying}, and secondly, where footprints deviated from the fitted power curve of GEDI L2A RH98 and PALSAR HV by more than 2.5 in decibels. 

\subsubsection{Sentinel-1}
We accessed and processed C-band Sentinel-1 data using GEE. Sentinel-1 ground range detected (GRD) images in the Interferometric Wide Swath (IW) mode, recorded during ascending orbits, were used in this study. These images were captured in dual-polarization, encompassing both the VV (Vertical Transmit, Vertical Receive) and VH (Vertical Transmit, Horizontal Receive) polarizations. Sentinel-1 GRD data have been subjected to essential preprocessing steps including thermal noise removal, radiometric calibration, and terrain correction, offering a ready-to-use dataset in decibels. We applied a focal mean filter to the VV and VH polarizations to further mitigate noise and enhance data clarity. This filtering approach aids in suppressing extremes in SAR data \citep{bonafilia2020sen1floods11}. We subsequently used the median values of the imagery acquired in 2019. Additionally, we incorporated the ratio of VV to VH polarizations into our analysis. The ratio serves as a more sensitive indicator for detecting changes in vegetation, owing to its reduced sensitivity to ground characteristics (e.g. soil moisture) \citep{veloso2017understanding, vreugdenhil2020sentinel}.

\subsubsection{Sentinel-2}
Sentinel-2 MultiSpectral Instrument (MSI) L2A data, acquired in 2019, were used in this study. Sentinel-2 L2A products provide orthorectified surface reflectance images, which have been corrected for atmospheric effects. The MSI sensor collects high-resolution optical images at 13 spectral bands spanning from the visible and near-infrared to the shortwave infrared, with spatial resolutions of 10 m, 20 m, and 60 m \citep{drusch2012sentinel}. We resampled all the bands to 10 m resolution using bicubic in GEE.To mitigate the effects of clouds, pre-filtering is initially applied to the Sentinel-2 images to retrieve granules with less than 10\% cloud cover. Subsequently, the QA60 band is used to remove pixels affected by clouds and cirrus \citep{wang2020mapping}. The 5-day revisit time of the combined constellation ensures that sufficient data remain after cloud removal. We utilized the median values of all the bands, as well as the Normalized Difference Vegetation Index (NDVI), kernel NDVI (kNDVI), and Normalized Difference Moisture Index (NDMI). The Normalized Difference Vegetation Index (NDVI) is sensitive to the greenness of vegetation \citep{rouse1974monitoring}. In addition, we derived annual minimum NDVI, maximum NDVI, and the difference between them for each year. Minimum NDVI represents base greenness in low vegetation activity period, while maximum NDVI indicates peak greenness and helps to reduce the effects of the atmospheric \citep{martinez2009vegetation}. Their difference quantifies annual vegetation dynamism, which helps account for vegetation seasonal variability when estimating forest biomass across time series. kNDVI enhances sensitivity in dense canopy \citep{camps2021unified}, and NDMI is responsive to vegetation water content \citep{wilson2002detection}. The multi-dimensional vegetation signals from NDVI, kNDVI and NDMI complement the optical bands, improving the representation of forest properties across different ecological conditions.

\subsubsection{ALOS PALSAR-2}
As an L-band SAR system, ALOS-2 PALSAR-2 can penetrate vegetation canopies and interact with woody components \citep{ni2014penetration}, providing sensitivity to forest structure beneath the canopy layer. The L-band ALOS-2 PALSAR-2 mosaic data acquired in 2019 is directly accessible via GEE. The 25m PALSAR-2 yearly mosaic was created by merging strips of PALSAR-2 imagery, and has been ortho-rectificatied and slope corrected by JAXA \citep{shimada2014new}. For each annual mosaic, strip data were chosen based on visual inspection, prioritizing images with minimal surface moisture response, and exclusively using data from that specific year without resorting to prior years for gap-filling (Figure x). We also applied a focal mean filter to the HV and HH polarizations to reduce noise. This step was crucial as the presence of speckle can induce significant uncertainties in AGB change detection \citep{mermoz2016forest}. The digital numbers (DN) of HV and HH were converted into gamma naught values in decibel using the following equation:
\begin{equation}
\gamma_0 = 10\log_{10}(\text{DN}^2) + CF \, 
\label{eq:palsar}
\end{equation}
where, $\gamma_0$ is the backscatter coefficient in dB, CF is the calibration factor equals to -83.0dB \citep{shimada2009palsar}. We incorporated the PALSAR-2 HV/HH ratio, which helps improve saturation point of AGB \citep{sarker2012potential}. In addition,  the local incidence angle was integrated to mitigate the effects of terrain \citep{zhang2019forest,das2015topographic}. Lastly, we resampled PALSAR-2 to a 10m resolution using bicubic interpolation to ensure compatibility with other datasets.

\subsubsection{Ancillary data}
In addition to the aforementioned remote sensing data, we also utilized topographic data and latitude-longitude data in this study. The topographic data, including elevation and slope derived from NASADEM \citep{jpl2020nasadem}, were incorporated to account for the variations in landscape and terrain, which significantly influence forest species and biomass accumulation \citep{stage2007interactions}. Latitude and longitude raster images were generated to represent the geographic coordinates of each pixel. Integrating longitude and latitude raster images can enrich models like RF by providing spatial context, which can potentially enhance the model's ability to capture and leverage spatial patterns and dependencies within the data, thereby improving prediction accuracy and model robustness. In addition, these latitude-longitude grids also provide additional geospatial contextual information beyond the patch scale to the deep learning models.

\subsubsection{Remote sensing data processing for machine learning models}
\begin{figure}[H]
    \centering
    \begin{subfigure}{1.0\textwidth}
        \centering
        \begin{tikzpicture}
            \node[anchor=south west,inner sep=0] (image) at (0,0) {\includegraphics[width=\textwidth]{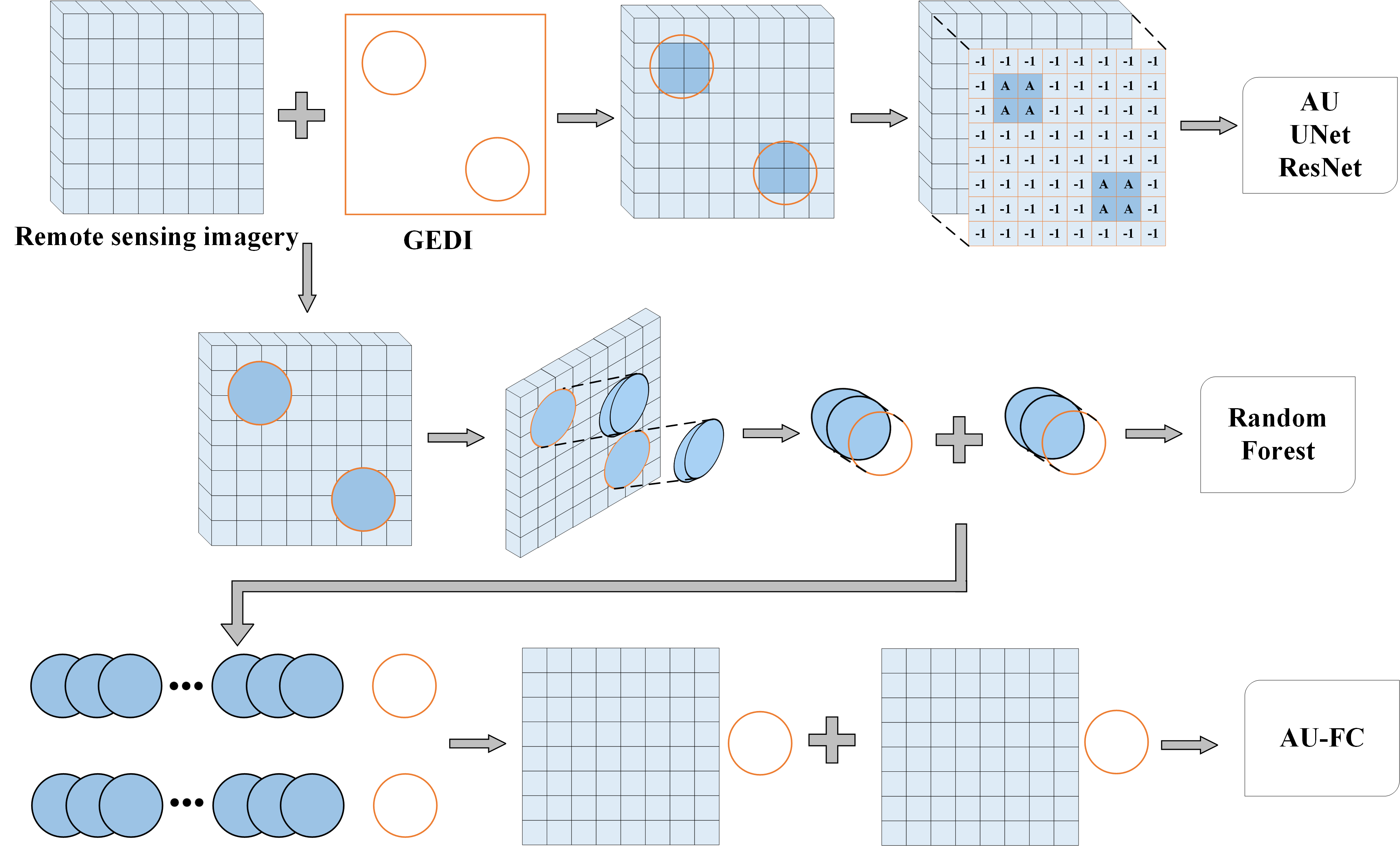}};
        \end{tikzpicture}
    \end{subfigure}%
\caption{Illustration of data processing for deep learning models and RF. A represents AGB derived from GEDI RH metrics}
\label{fig: Data processing}
\end{figure}

In order to estimate AGB using remote sensing data through deep learning and RF algorithms, it is imperative to transform both remote sensing imagery and footprint data into a format suitable for model input. Specifically, the initial sets of remote sensing imagery were segmented into image patches, each with dimensions of 64 pixels by 64 pixels and comprising 29 bands. Subsequently, GEDI footprints were employed to construct labels for the deep learning model. If the centroid of a pixel lies within a GEDI footprint, its label is assigned the value of AGB derived from GEDI; otherwise, it is set to -1 (Fig. \ref{fig: Data processing}). For the RF model, the mean value of pixels within the GEDI footprints was extracted using an area-weighted method. Partially intersected pixels along polygon boundaries contribute proportionally based on their overlapped area rather than fully counted. Then the mean value of each GEDI footprint was transformed to a patch used for a deep learning model without spatial information.

\subsection{Forest AGB estimation}
\begin{figure}[H]
    \centering
    \begin{subfigure}{1.0\textwidth}
        \centering
        \begin{tikzpicture}
            \node[anchor=south west,inner sep=0] (image) at (0,0) {\includegraphics[width=\textwidth]{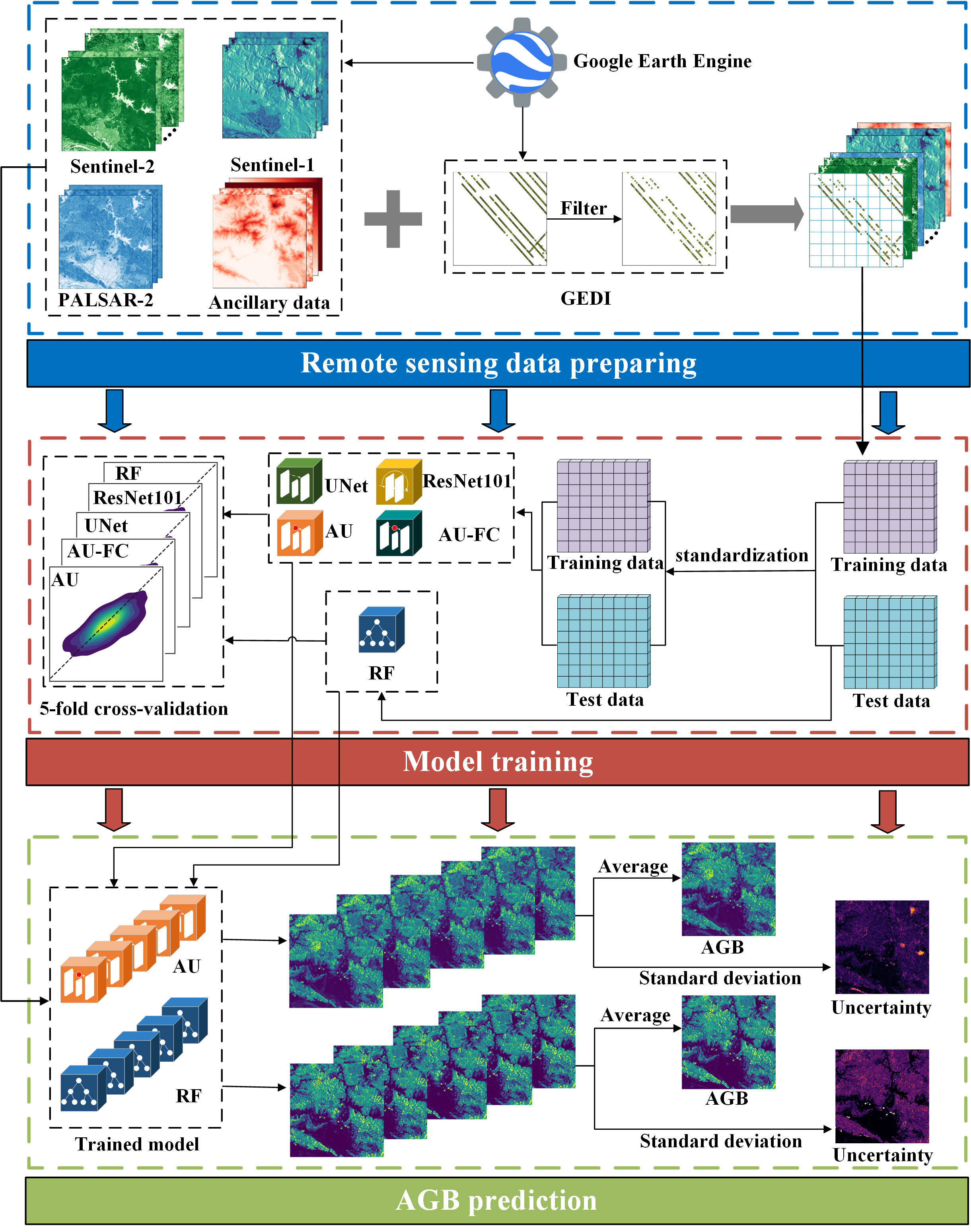}};
        \end{tikzpicture}
    \end{subfigure}%
\caption{General framework.}
\label{fig: Framework}
\end{figure}

We developed a new approach to estimate forest AGB using a combination of satellite images and deep learning algorithms. The AGB derived from GEDI using the following equation were used as the label. 
\begin{equation}
\text{AGB} = 5.58 \text{RH80}^{1.12}
\end{equation}
All remote sensing images were resampled to a 10m resolution using bicubic interpolation to extrapolate AGB derived from GEDI. In this study, we employed a 5-fold cross-validation approach \citep{yu2022estimation} to compare the accuracies of the deep learning models and RF, and utilized both AU and RF to generate forest AGB maps for Guangdong, China (Fig. \ref{fig: Framework}). This involved conducting five distinct computations during the model accuracy comparison, utilizing the scatter plots with the highest precision for each model. Five AGB maps were generated using the trained models and subsequently, the mean of these five iterations was utilized as the final AGB map to obtain a more robust and reliable AGB estimation. In addition, an uncertainty map was derived by calculating the standard deviation of the 5 AGB predictions at each pixel location. Both our AGB and uncertainty maps were masked using a forest cover map. The forest cover map was extracted from the European Space Agency (ESA) WorldCover 10m v200 \citep{zanaga2022esa}.

\subsubsection{Attention UNet}
In this study, we aim to establish an end-to-end nonlinear mapping between remote sensing data and the corresponding AGB using the attention-based deep learning network. The mathematical formulation of this objective is as follows,
\begin{equation}
f=\arg \min _\theta \frac{1}{N} \sum_{i=1}^N\left\|f\left(X_i\right)-Y_i\right\|_2^2
\end{equation}
where \( f \) denotes the AU network model utilized in this study, \( f(X_i) \) signifies the estimated AGB, and \( Y_i \) refers to the AGB values derived from GEDI data. Additionally, \( N \) represents the total number of samples present in the training dataset.

\begin{figure}[H]
    \centering
    \begin{subfigure}{1.0\textwidth}
        \centering
        \begin{tikzpicture}
            \node[anchor=south west,inner sep=0] (image) at (0,0) {\includegraphics[width=\textwidth]{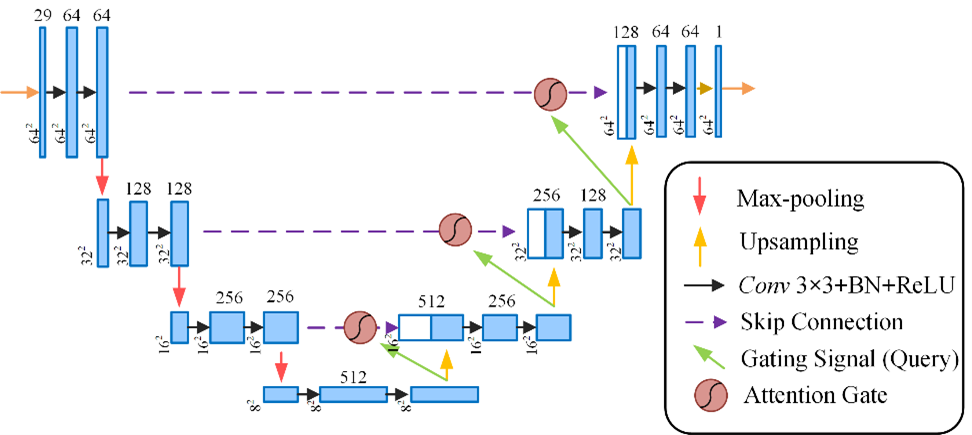}};
        \end{tikzpicture}
    \end{subfigure}%
\caption{The architecture of AU.}
\label{fig: Attention UNet}
\end{figure}

As illustrated in fig. \ref{fig: Attention UNet}, the architecture of the proposed AU model is fundamentally built upon the UNet framework. The model accepts remote sensing data with dimensions of 64×64×29 as input. The architecture is modular, consisting of both encoding and decoding stages, each containing three structurally identical blocks. In each encoding block, the input data undergo two successive 3×3 convolution layers, each accompanied by batch normalization (BN) and a rectified linear unit (ReLU) activation function, before proceeding to a 2×2 max-pooling layer for spatial down-sampling. Following the encoding blocks, the bottom layer is subjected to up-sampling via a 2×2 kernel. The decoding blocks mirror their encoding counterparts, also featuring dual 3×3 convolution layers with batch normalization and ReLU activations. In addition, feature maps extracted at multiple scales are integrated through skip connections. This integration serves to combine coarse- and fine-level features from the remote sensing data. The architecture concludes with a 1×1 convolutional kernel to fine-tune the channel dimensions of the final output. As a result, estimated AGB values with dimensions of 64×64 are obtained.

\begin{figure}[H]
    \centering
    \begin{subfigure}{1.0\textwidth}
        \centering
        \begin{tikzpicture}
            \node[anchor=south west,inner sep=0] (image) at (0,0) {\includegraphics[width=\textwidth]{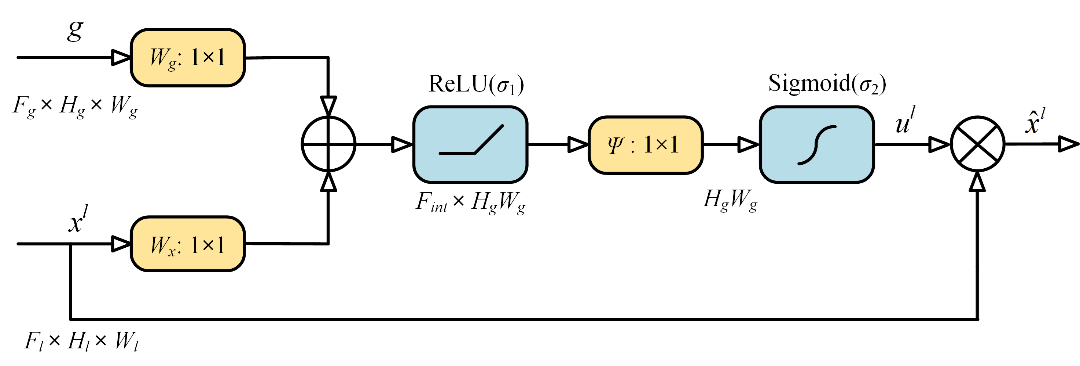}};
        \end{tikzpicture}
    \end{subfigure}%
\caption{The structure of the attention gate.}
\label{fig: Attention gate}
\end{figure}

Extracting features that are strongly associated with AGB is a non-trivial task. To address this, our architecture incorporates an additive spatial self-attention mechanism in the skip connections. This modification enables the model to selectively emphasize salient features while attenuating less relevant features during the training process. By directing the focus of the model in this manner, we substantially enhance its predictive accuracy for AGB. Moreover, the integration of attention gates into the UNet framework introduces negligible computational overhead while delivering a notable improvement in both model sensitivity and accuracy. As depicted in Fig. \ref{fig: Attention gate}, the implementation of the additive attention gate within the skip-connection segment is elaborated. 
Here, F denotes the number of feature maps, while H and W represent the dimensions in terms of height and width of the feature maps, respectively. The mathematical formulation of the attention gate is provided below,
\begin{equation}
q_{a t t}^l=\psi^T\left(f_1\left(W_x^T x_i^l+W_g^T g_i+b_x+b_g\right)\right)+b_\psi
\end{equation}
where \( W_x \) and \( W_g \) serve as the matrices for feature weighting, \( x_i^l \in \mathbb{R}^{F_l} \) and \( g_i \in \mathbb{R}^{F_g} \) correspond to the encoding matrix and the decoding matrix, respectively. The activation function \( f_1 \) is implemented using ReLU. Additionally, \( \psi \) represents a convolutional operator that employs a \( 1 \times 1 \) kernel. The bias terms \( b_x \), \( b_g \) and \( b_\psi \) are affiliated with their respective convolutional layers during the training process. Lastly, \( q_{a t t}^l \) functions as an intermediate representation in the computation. In the attention gate architecture, features extracted from coarser scales are used in gating to discern and eliminate irrelevant or noisy responses in the skip connections. This adaptive filtering mechanism is active during both the forward and backward propagation phases, thereby optimizing neuron activations and contributing to enhanced model performance.

The attention coefficient is formulated as given below,
\begin{equation}
u_i^l=f_2\left(q_{a t t}^l\left(x_i^l, g_i ; \Theta_{a t t}\left(W_x, W_g, \psi, b_x, b_g, b_\psi\right)\right)\right)
\end{equation}
where \( f_2 \) serves as the Sigmoid activation function, while $\Theta_{a t t}$ represents a set of parameters containing linear transformations \( W_x \in \mathbb{R}^{F_l \times F_{\text{int}}} \), \( W_g \in \mathbb{R}^{F_g \times F_{\text{int}}} \), \( \psi \in \mathbb{R}^{F_{\text{int}} \times 1} \), and bias terms \( b_x \in \mathbb{R}^{F_{\text{int}}} \), \( b_g \in \mathbb{R}^{F_{\text{int}}} \), \( b_{\psi} \in \mathbb{R} \).

The output of the attention gate, denoted as \( \hat{x}_i^l \), is obtained through element-wise multiplication between the attention coefficient \( u_i^l \) and the encoding matrix \( x_i^l \). As delineated in (\ref{eq: The output of the attention gate}), this computational step amplifies the saliency of features that exhibit a robust correlation with AGB, while effectively attenuating the contributions from extraneous or less pertinent features.
\begin{equation}
 \hat{x}_i^l  = u_i^l * x_i^l
 \label{eq: The output of the attention gate}
\end{equation} 
The update formulation for the convolutional parameters at layer $l-1$ is presented as follows:
\begin{equation}
\frac{\partial\left(\hat{x}_{i}^{l}\right)}{\partial\left(\Phi^{l-1}\right)}=\frac{\partial\left(u_{i}^{l} f\left(x_{i}^{l-1} ; \Phi^{l-1}\right)\right)}{\partial\left(\Phi^{l-1}\right)}=u_{i}^{l} \frac{\partial\left(f\left(x_{i}^{l-1} ; \Phi^{l-1}\right)\right)}{\partial\left(\Phi^{l-1}\right)}+\frac{\partial\left(u_{i}^{l}\right)}{\partial\left(\Phi^{l-1}\right)} x_{i}^{l}
\label{eq: partial derivatives}
\end{equation}

\subsubsection{Fully connected AU}
The attention UNet fully connected (AU-FC) model was employed to validate the performance of AU in scenarios where spatial information is not utilized. This model assesses the capability of AU to effectively discern and learn pertinent features and patterns without leveraging the inherent spatial relationships within the remote sensing data. The input data for this model, based on the input data utilized for the RF, was transformed via a fully connected layer, feeding into 4096 nodes and then reshaping the nodes to 64*64 patches suitable for processing by the AU model (Fig. \ref{fig: Data processing}). We employed the fully connected layer to connect every input data to every output data \citep{liu2018breast}, and the fully connected layer was denoted as follows:
\begin{equation}
 y = fc(x, w, b)
 \label{eq: The fully connected layer}
\end{equation}
where \(x\) is the vector representing the input to a fully connected layer, \(w\) is the corresponding weight matrix, and \(b\) is the bias vector. The output of this layer is represented by the vector \(y\). The input vector \(x\) encompasses \(M\) distinct elements, with each element denoted as \(x_i\), where \(i\) represents the index of the element. It is crucial to note that the weight matrix \(w\) has dimension of \(M \times K\), with \(K\) symbolizing the dimension of the output vector \(y\). This layer operates by transforming the input vector \(x\), subsequently generating each element, \(y_{i'}\), in the output vector \(y\), according to  to the following function.
\begin{equation}
 y_{i^{\prime}}=\sum_{i} w_{i i^{\prime}} x_{i}+b_{i^{\prime}}
 \label{eq: Each output of the fully connected layer}
\end{equation}
Subsequent to the predictions made by the AU model, the data were transformed back through two additional fully connected layers into discrete AGB predictions. The first fully connected layer condensed the spatial encoding into a 2048-vector embedding. The second layer mapped this to a single AGB value per sample. The fully connected layer essentially functions as a projection to align the inputs and outputs.

\subsubsection{ResNet101}
In this study, the deep residual network (ResNet) was also employed to estimate forest biomass. This ResNets introduce residual learning, where the models learn residual functions with reference to the layer inputs, which enables the training of very deep networks by allowing the flow of gradients through the network during backpropagation \citep{he2016deep}. The core idea of ResNets is to add skip connections that bypass some layers, with their outputs being directly added to later layers. This helps gradients propagate through the network during backpropagation, mitigating the vanishing gradient problem associated with training deep neural networks especially as the depth increases. The extremely deep architecture of ResNet101, with over 100 layers, provides high modeling capacity to extract features at multiple levels of abstraction \citep{xu2020research}. We thus select ResNet101 to estimate forest AGB based on its powerful representational abilities

\subsubsection{Model setting}\label{sec: Model setting}
A. Loss function

The task of predicting AGB is cast as a regression problem, for which the loss function utilized in our model is articulated as follows:
\begin{equation}
\mathrm{L}_{\text {loss }}=\mathrm{L}_{M S E}+\lambda\|\theta\|^2
\end{equation}
In this equation, the first term represents the Mean Squared Error (MSE) loss component, while the second term denotes the \( l_2 \) regularization term with a regularization parameter \( \lambda \). \( \theta = \{W, b\} \) denotes the training parameters: weights and bias of the network.

B. Data preprocessing

The image patches were partitioned at random into three distinct datasets: a training set, a validation set, and a test set, following a ratio of 7:2:1, respectively \citep{yu2022multiscale}. The training set comprises 39,116 samples, while the validation and test sets contain 1,1176 and 5,590 samples, respectively. Prior to being fed into the neural networks, both the remote sensing data and the associated AGB undergo standardization. Standardization significantly enhances the ability of network to generalize across different datasets. The standardization process is mathematically represented by (\ref{eq: standardization}),

\begin{equation}
D_i^{\prime} = \frac{D_i - D_{\text{{mean\_train}}}}{D_{\text{{std\_train}}}}
\label{eq: standardization}
\end{equation}

where $D_i$ denotes either the remote sensing data or labels and $D_i'$ represents the corresponding normalized data. $D_{\text{{mean\_train}}}$ and $D_{\text{{std\_train}}}$ signify the mean and standard deviation for each channel in the training set, respectively. It should be noted that the input remote sensing data and labels in validation and test sets are also normalized using $D_{\text{{mean\_train}}}$ and $D_{\text{{std\_train}}}$, respectively. This strategy is employed to prevent data leakage and to ensure a robust evaluation of the generalization performance of the model.

C. Implementation details and computational configuration

The deep learning algorithms were executed on a dedicated server, configured with an NVIDIA RTX A5000 24GB GPU, an Intel\textsuperscript{\textregistered} Xeon\textsuperscript{\textregistered} W-2265 CPU (3.5 GHz), and 128 GB of RAM. The model was developed in Python 3.10 and utilized PyTorch 2.0.1 for backend computations. The Adam optimizer \citep{kingma2014adam} was employed in the deep learning methods. Furthermore, the initial learning rate was configured to 0.001 and was designed to decay by a factor of $0.1$ at every 40th epoch. The training process was constrained to a maximum of 120 epochs. A batch size of 128 was employed, and an \(L_2\) regularization term for weight decay was set to $1 \times 10^{-5}$ as a countermeasure against overfitting. In accordance with the deep supervision strategy, we employ a masking technique on regions devoid of labels during the loss computation, thereby addressing the inherent sparsity of the ground-truth labels.

\subsubsection{Mitigating boundary effects in patch-based predictions} \label{sec: Mitigating boundary effects}
UNet and other deep learning models frequently demonstrate diminished accuracy and reliability at the boundary or edge regions of images \citep{innamorati2020learning}, a phenomenon often referred to as the "boundary effect" or "edge effect". In addition, independent per-patch inference during prediction leads to spatial incoherence and discontinuities between adjacent patches predicted by the deep learning models. This is predominantly because each patch is processed and evaluated independently, without considering the spatial context and relational information from its neighboring patches. To mitigate the boundary effect, a strategy was deployed during the prediction phase where each patch has an overlap of 10 pixels with its neighboring patches. Subsequently, the outermost 3 pixels of each patch were discarded to alleviate the influence of potentially inaccurate predictions at the patch boundaries. The overlapping regions were then averaged to obtain a smooth transition between patches. This approach aims to enhance spatial coherence between neighboring patches and refine the model's predictive accuracy at the boundary regions, thus providing more reliable and consistent output in image analysis tasks.

\subsubsection{Accuracy assessment}
To quantitatively evaluate the predictive performance of various methods on AGB, three regression metrics are employed: the coefficient of determination, commonly known as R-squared (R$^2$), the root mean square error (RMSE), and the bias. In the regression problem, these three metrics convincingly reflect the predictive capability of methods, and the definitions are as follows,

\begin{equation}
\mathrm{R}^2=1-\frac{S S_{\mathrm{res}}}{S S_{\text {tot }}}=1-\frac{\sum_i^K\left(Y_i-\hat{Y}_i\right)^2}{\sum_i^K\left(Y_i-\bar{Y}_i\right)^2}
\end{equation}
\begin{equation}
\mathrm{RMSE}=\sqrt{\frac{1}{K} \sum_{i=1}^K\left(Y_i-\hat{Y}_i\right)^2} 
\end{equation}
\begin{equation}
\mathrm{Bias} = \frac{1}{K} \sum_{i=1}^{K} (\hat{Y}_i - Y_i)
\end{equation}

where \( K \) denotes the cumulative number of observations, \( Y_i \) is the AGB derived from GEDI, \( \hat{Y}_i \) denotes the predicted AGB, and \( \bar{Y} \) is the mean of the AGB derived from GEDI.

\section{Results}
\subsubsection{Model performance}
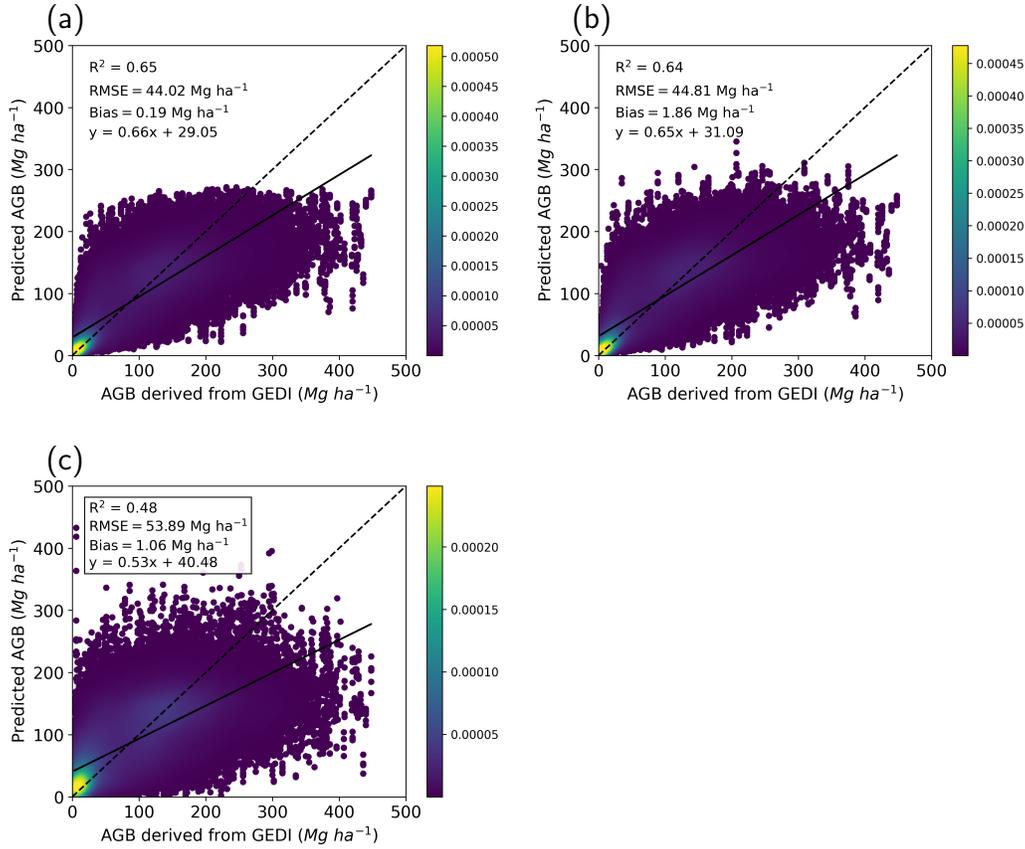
\begin{figure}[H]
    \centering
    \begin{subfigure}{0.49\textwidth}
        \centering
        \begin{tikzpicture}
            \node[anchor=south west,inner sep=0] (image) at (0,0) {\includegraphics[width=\textwidth]{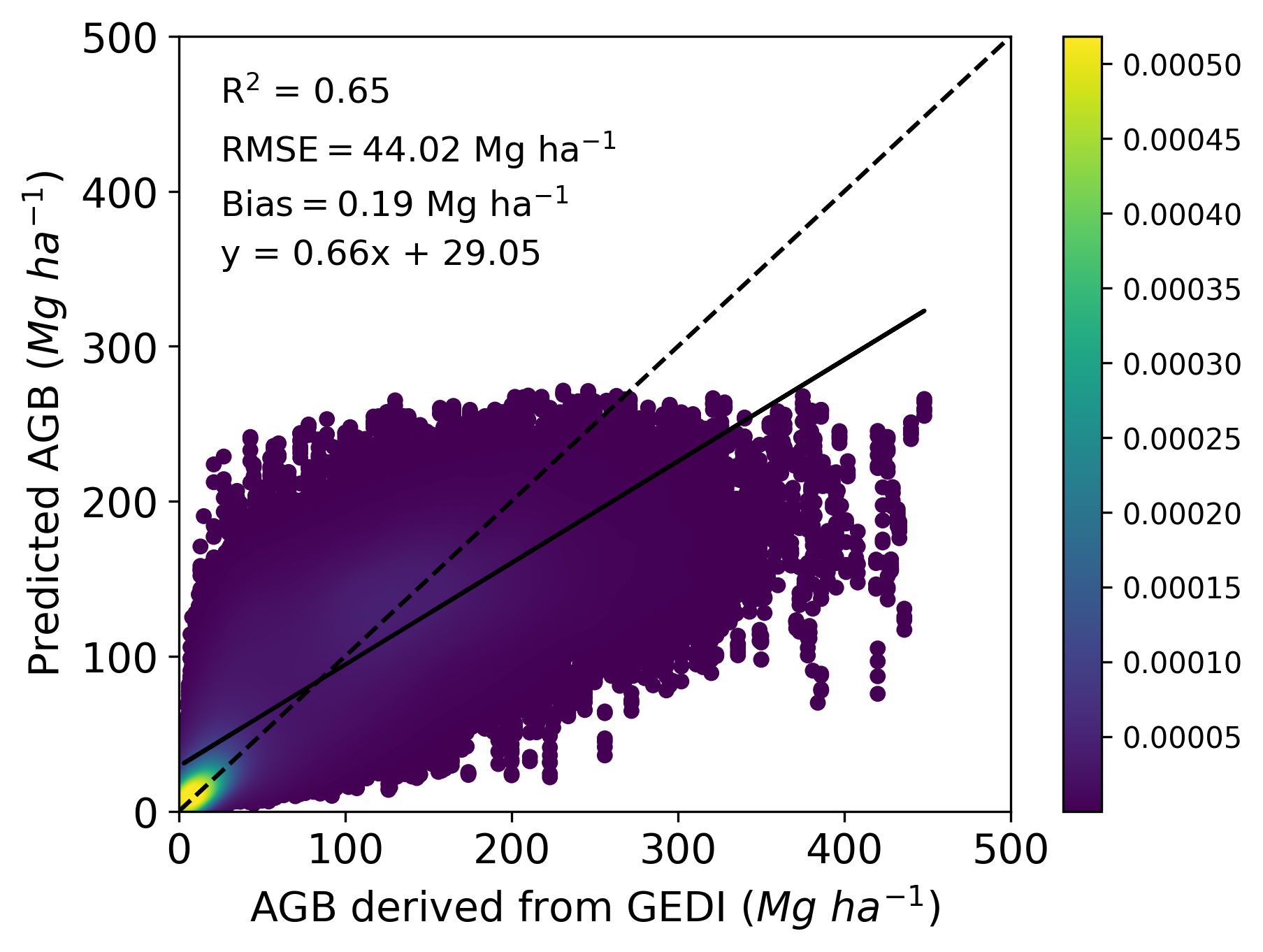}};
            \node[rectangle, draw=none, fill=white, inner sep=0pt, anchor=north west, font=\sffamily, xshift=17pt, yshift=10pt] at (image.north west) {(a)};
        \end{tikzpicture}
    \end{subfigure}%
    \hfill
    \begin{subfigure}{0.49\textwidth}
        \centering
        \begin{tikzpicture}
            \node[anchor=south west,inner sep=0] (image) at (0,0) {\includegraphics[width=\textwidth]{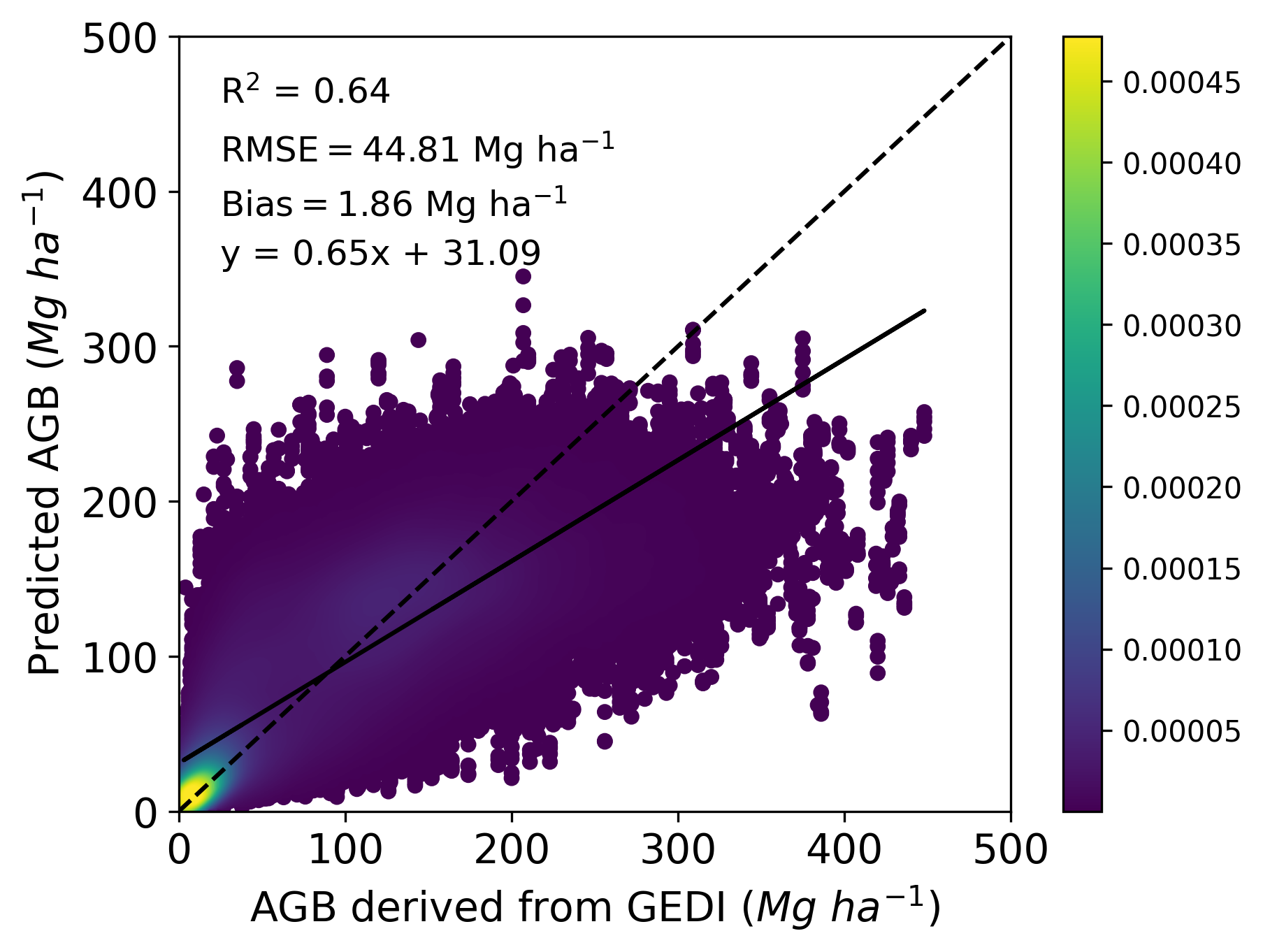}};
            \node[rectangle, draw=none, fill=white, inner sep=0pt, anchor=north west, font=\sffamily, xshift=17pt, yshift=10pt] at (image.north west) {(b)};
        \end{tikzpicture}
    \end{subfigure}%
    
    \vspace*{\floatsep}
    
    \begin{subfigure}{0.49\textwidth}
        \centering
        \begin{tikzpicture}
            \node[anchor=south west,inner sep=0] (image) at (0,0) {\includegraphics[width=\textwidth]{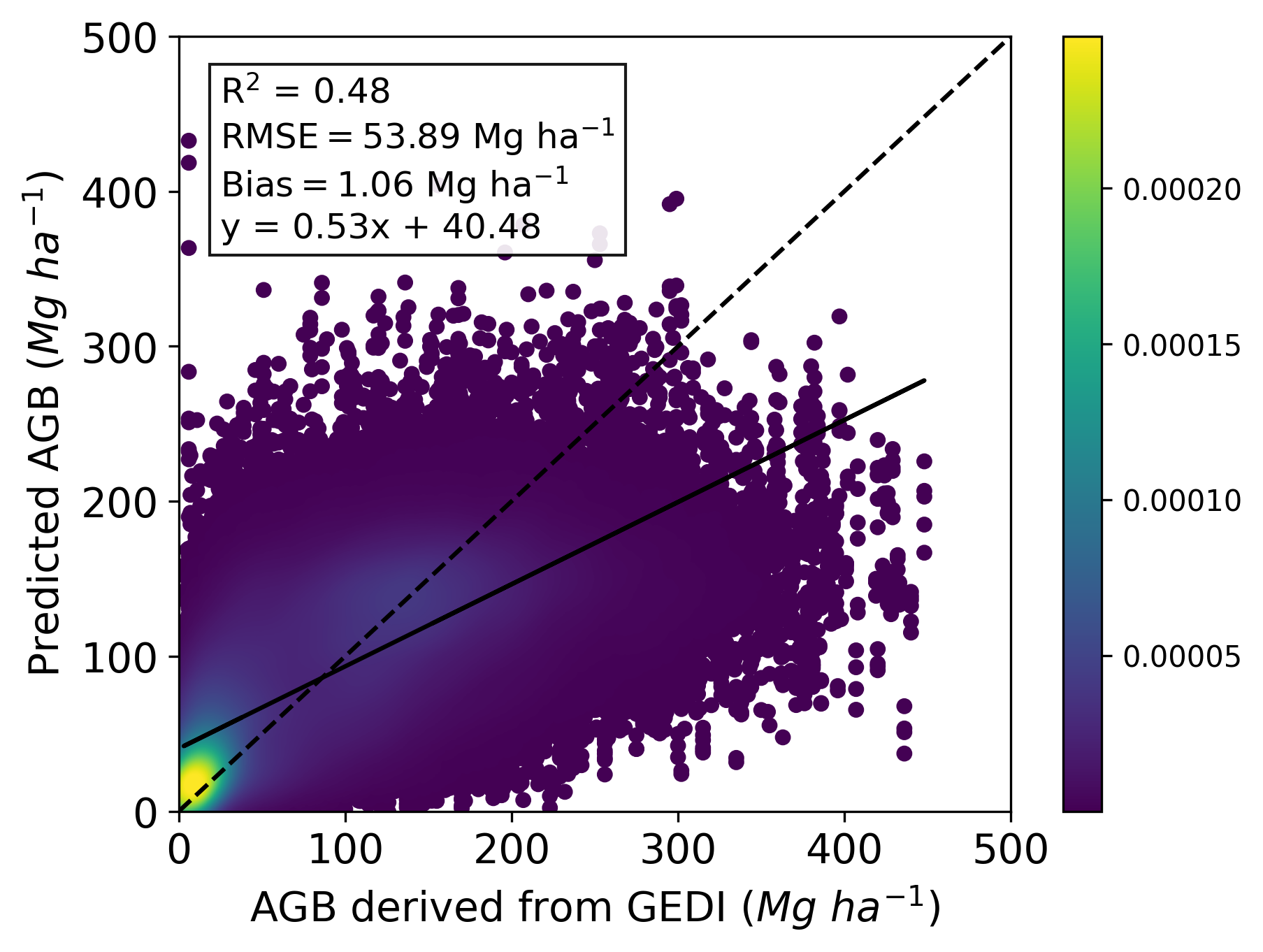}};
            \node[rectangle, draw=none, fill=white, inner sep=0pt, anchor=north west, font=\sffamily, xshift=17pt, yshift=10pt] at (image.north west) {(c)};
        \end{tikzpicture}
    \end{subfigure}%
    \hfill
    \begin{subfigure}{0.49\textwidth}
        \centering
    \end{subfigure}
    
    \caption{The predicted AGB values against AGB derived from GEDI RH metrics. (a) AU, (b) UNet, (c) ResNet101. The color represents the estimated density of points at that specific location.}
    \label{fig: model performance}
\end{figure}

To evaluate the accuracy and reliability of the deep learning models in estimating AGB, we conducted five separate AGB estimations for AU \citep{oktay2018attention}, UNet \citep{ronneberger2015u} and ResNet \citep{he2016deep}. For each time, we employed unique combinations of training and validation sets, ensuring a comprehensive understanding of the models' adaptability and consistency. The performance of the three deep learning methods across the five estimations remained consistent when evaluated using the test dataset, with the highest accuracy instance shown in Fig. \ref{fig: model performance}. Based on the metrics of R$^2$, RMSE, and bias, AU clearly outperforms the other two models. Additionally, from the scatter plot, it is evident that AU has fewer outliers compared to UNet and ResNet101, further underscoring its reliability in AGB estimation. The R$^2$ values for AU and UNet are 0.65 and 0.64, respectively, and their RMSE values are closely matched at 44.02 Mg ha\textsuperscript{-1} and 44.81 Mg ha\textsuperscript{-1}. Notably, AU's bias is 0.19 Mg ha\textsuperscript{-1}, which is considerably lower than UNet's bias, which stands at 1.86 Mg ha\textsuperscript{-1}. In contrast, ResNet101's performance is less impressive, with an R$^2$ of 0.48 and an RMSE of 53.89 Mg ha\textsuperscript{-1}. Its bias is 1.06 Mg ha\textsuperscript{-1}, placing it between AU and UNet.

\begin{figure}[H]
    \centering
    \begin{subfigure}{0.49\textwidth}
        \centering
        \begin{tikzpicture}
            \node[anchor=south west,inner sep=0] (image) at (0,0) {\includegraphics[width=\textwidth]{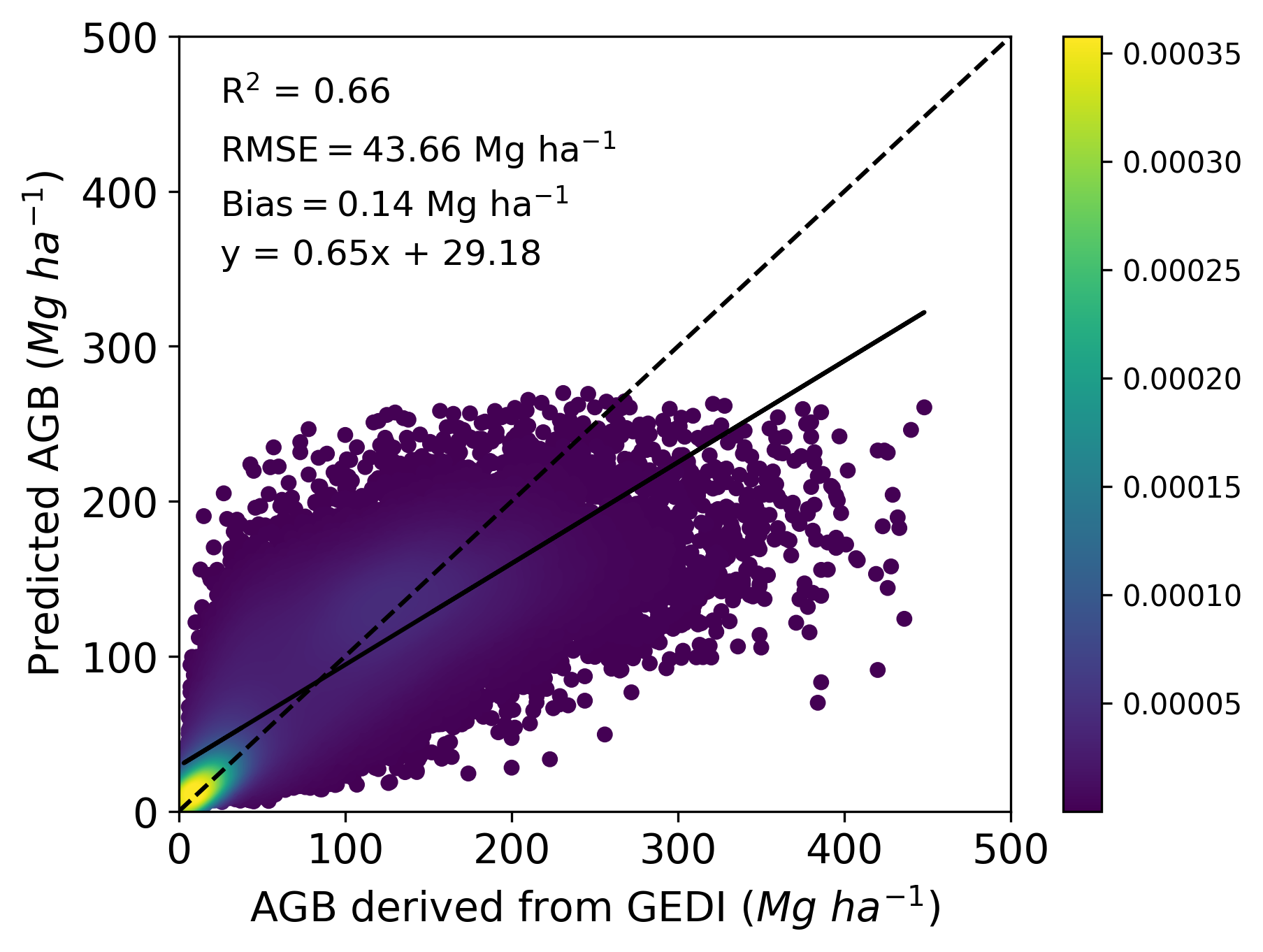}};
            \node[rectangle, draw=none, fill=white, inner sep=0pt, anchor=north west, font=\sffamily, xshift=17pt, yshift=10pt] at (image.north west) {(a)};
        \end{tikzpicture}
    \end{subfigure}%
    \hfill
    \begin{subfigure}{0.49\textwidth}
        \centering
        \begin{tikzpicture}
            \node[anchor=south west,inner sep=0] (image) at (0,0) {\includegraphics[width=\textwidth]{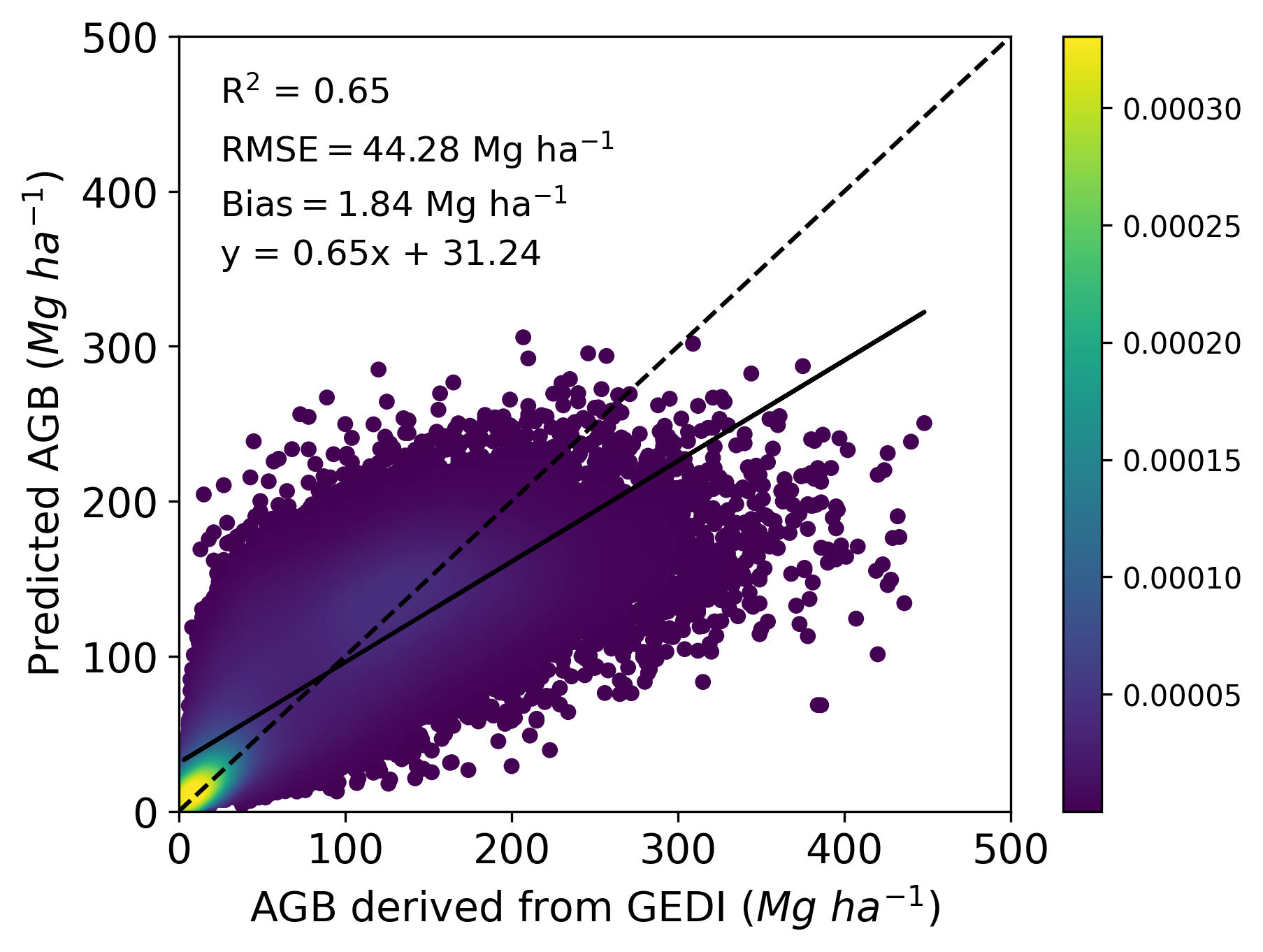}};
            \node[rectangle, draw=none, fill=white, inner sep=0pt, anchor=north west, font=\sffamily, xshift=17pt, yshift=10pt] at (image.north west) {(b)};
        \end{tikzpicture}
    \end{subfigure}%
    \vspace*{\floatsep}
    
    \begin{subfigure}{0.49\textwidth}
        \centering
        \begin{tikzpicture}
            \node[anchor=south west,inner sep=0] (image) at (0,0) {\includegraphics[width=\textwidth]{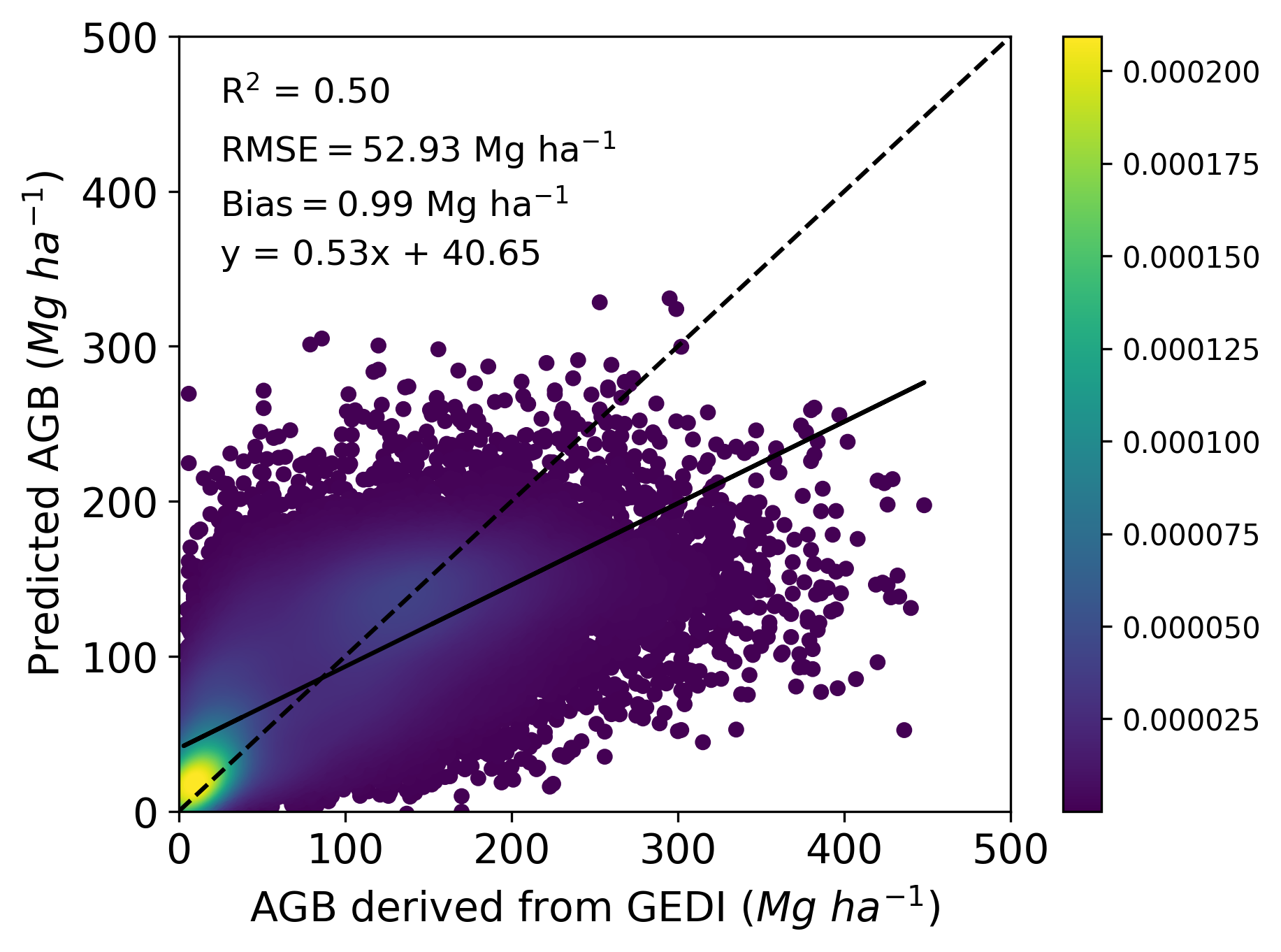}};
            \node[rectangle, draw=none, fill=white, inner sep=0pt, anchor=north west, font=\sffamily, xshift=17pt, yshift=10pt] at (image.north west) {(c)};
        \end{tikzpicture}
    \end{subfigure}%
    \hfill
    \begin{subfigure}{0.49\textwidth}
        \centering
        \begin{tikzpicture}
            \node[anchor=south west,inner sep=0] (image) at (0,0) {\includegraphics[width=\textwidth]{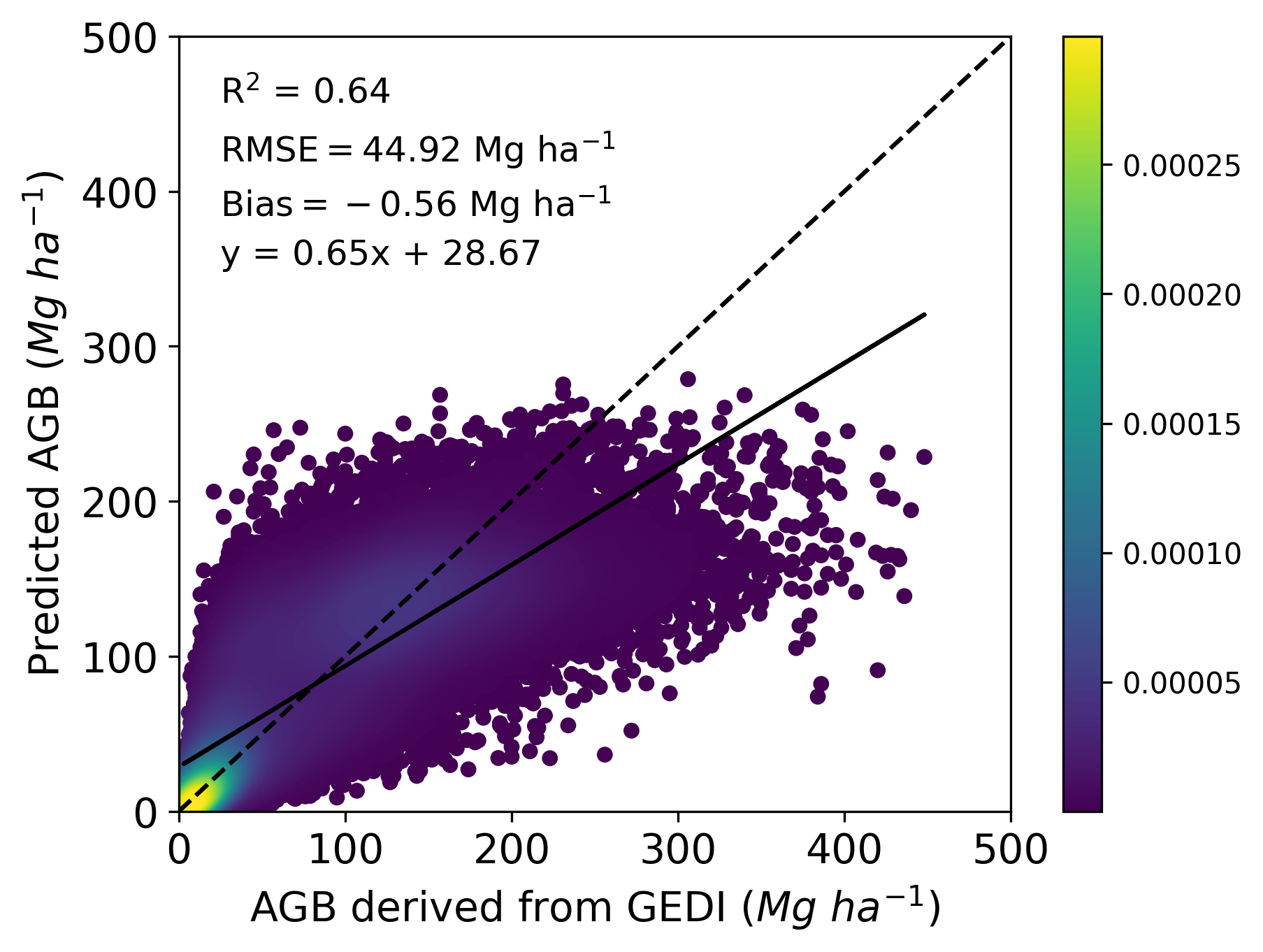}};
            \node[rectangle, draw=none, fill=white, inner sep=0pt, anchor=north west, font=\sffamily, xshift=17pt, yshift=10pt] at (image.north west) {(d)};
        \end{tikzpicture}
    \end{subfigure}%
    \vspace*{\floatsep}
    
    \begin{subfigure}{0.49\textwidth}
        \centering
        \begin{tikzpicture}
            \node[anchor=south west,inner sep=0] (image) at (0,0) {\includegraphics[width=\textwidth]{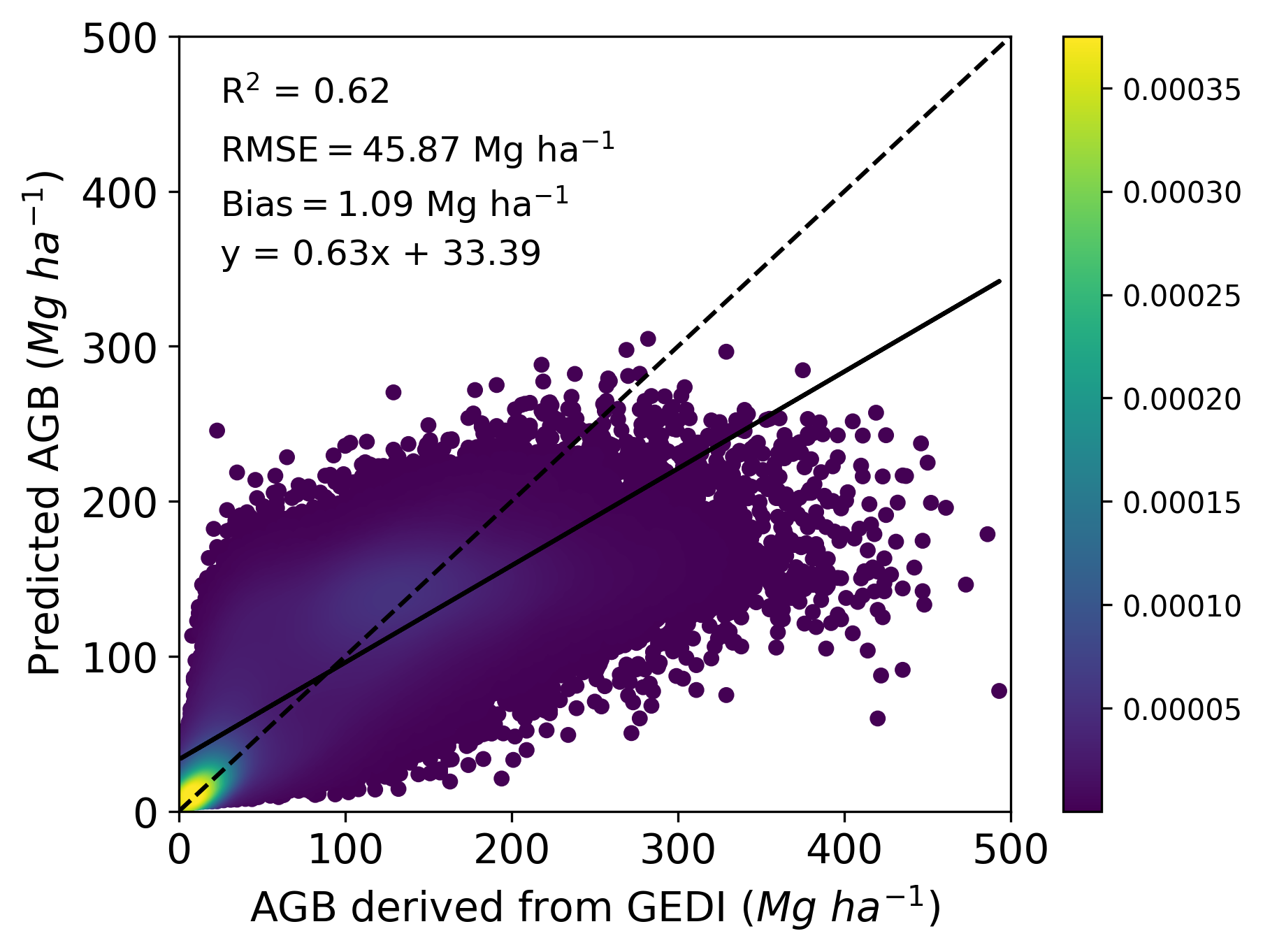}};
            \node[rectangle, draw=none, fill=white, inner sep=0pt, anchor=north west, font=\sffamily, xshift=17pt, yshift=10pt] at (image.north west) {(e)};
        \end{tikzpicture}
    \end{subfigure}%
    \hfill
    \begin{subfigure}{0.49\textwidth}
        \centering
    \end{subfigure}
    
    \caption{Mean predicted AGB values within GEDI footprints plotted against AGB derived from GEDI RH metrics. (a) AU, (b)UNet, (c) ResNet101, (d) AU-FC (e) RF.}
    \label{fig: model performance mean values}
\end{figure}
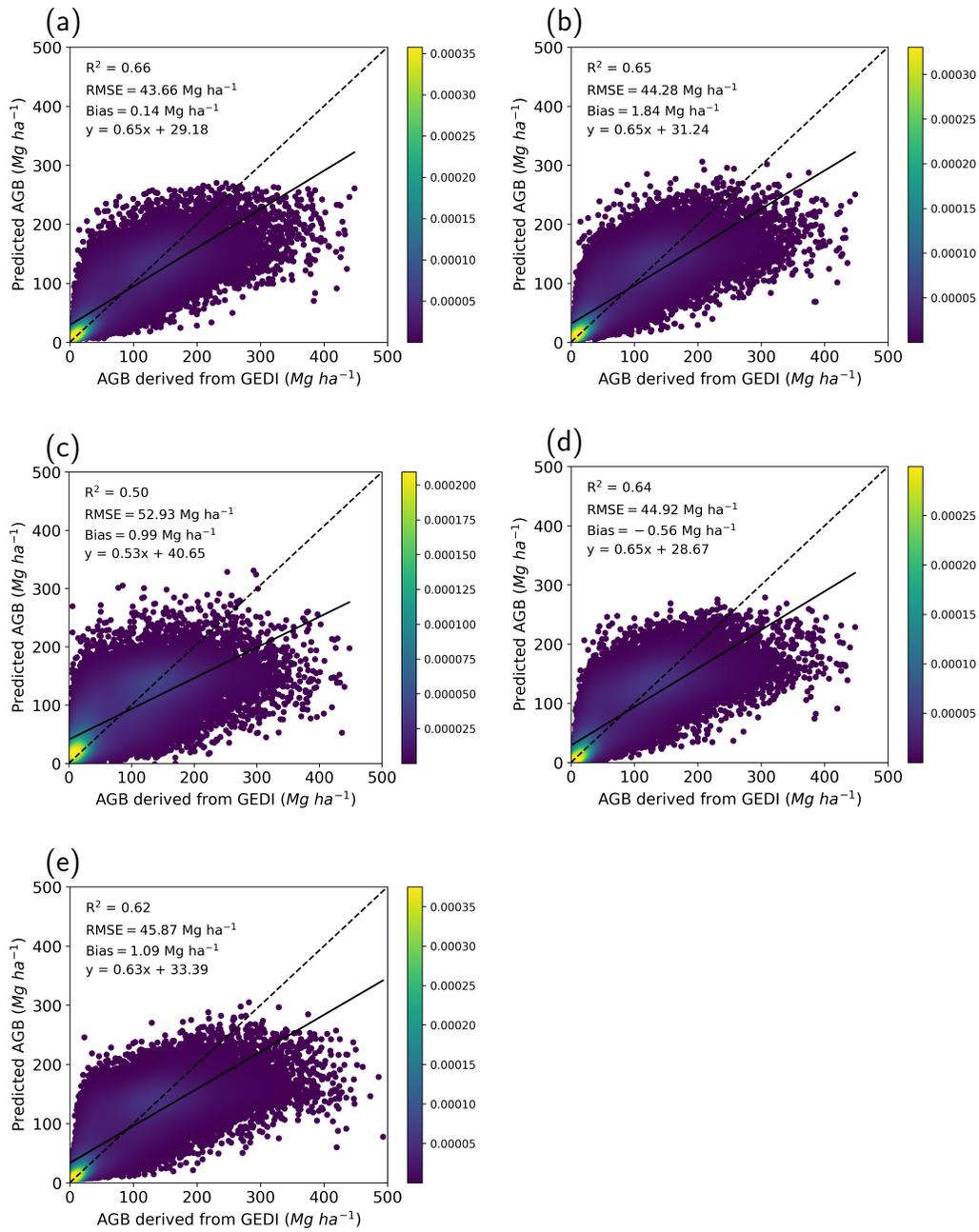

To further assess the performance of deep learning models, we compared the deep learning models with a commonly utilized traditional machine learning method, the RF. To enable comparison with the RF approach, the deep learning AGB estimates for each individual GEDI footprint were aggregated by computing the mean prediction within each footprint. Scatter plots were generated between these mean AGB predictions and the corresponding AGB derived from GEDI (Fig. \ref{fig: model performance mean values}). Compared to the approach of scattering the predictions from individual pixels, employing footprint-level means yields an increased R$^2$ and a decreased RMSE, as evidenced by a comparison of Fig. \ref{fig: model performance mean values} (a) (b) (c) with Fig. \ref{fig: model performance} (a) (b) (c). This enhancement can be attributed to the noise and variability attenuation achieved by averaging at the pixel level. By computing the mean, outliers and specific anomalies within each footprint are effectively smoothed out.

The performance of the AU model is markedly superior to that of the RF, as demonstrated by its higher R$^2$ value, coupled with reduced RMSE and bias metrics. Notably, the minimal bias of the AU model indicates that, on average, its predictions are closely aligned with the true values, without consistently under- or over-estimating. This superiority underscores AU's ability to capture underlying patterns and trends more effectively than traditional machine learning methods. To explore whether the enhanced accuracy of AU over RF was solely attributed to its utilization of spatial information, an additional experiment was conducted comparing AU-FC (Fig. \ref{fig: model performance mean values} (d)). The results revealed the prediction accuracy of AU-FC was intermediate between AU and RF models. This intermediate performance suggests that while the integration of spatial information by AU does contribute to its superior accuracy, other inherent characteristics of the model also play a significant role in optimizing prediction outcomes. It underscores the multifaceted nature of model accuracy, influenced not just by the incorporation of spatial context but also by the underlying architecture and learning mechanisms of the model. However, deep learning models are not necessarily superior to traditional machine learning algorithms. While the UNet boasts a higher R$^2$ and lower RMSE than RF, its bias is greater. Conversely, ResNet101 has a lower R$^2$ and higher RMSE compared to RF, with only its bias being marginally lower. 

\subsubsection{Enhancement of spatial consistency}
\begin{figure}[H]
    \centering
    \begin{subfigure}{1.0\textwidth}
        \centering
        \begin{tikzpicture}
            \node[anchor=south west,inner sep=0] (image) at (0,0) {\includegraphics[width=\textwidth]{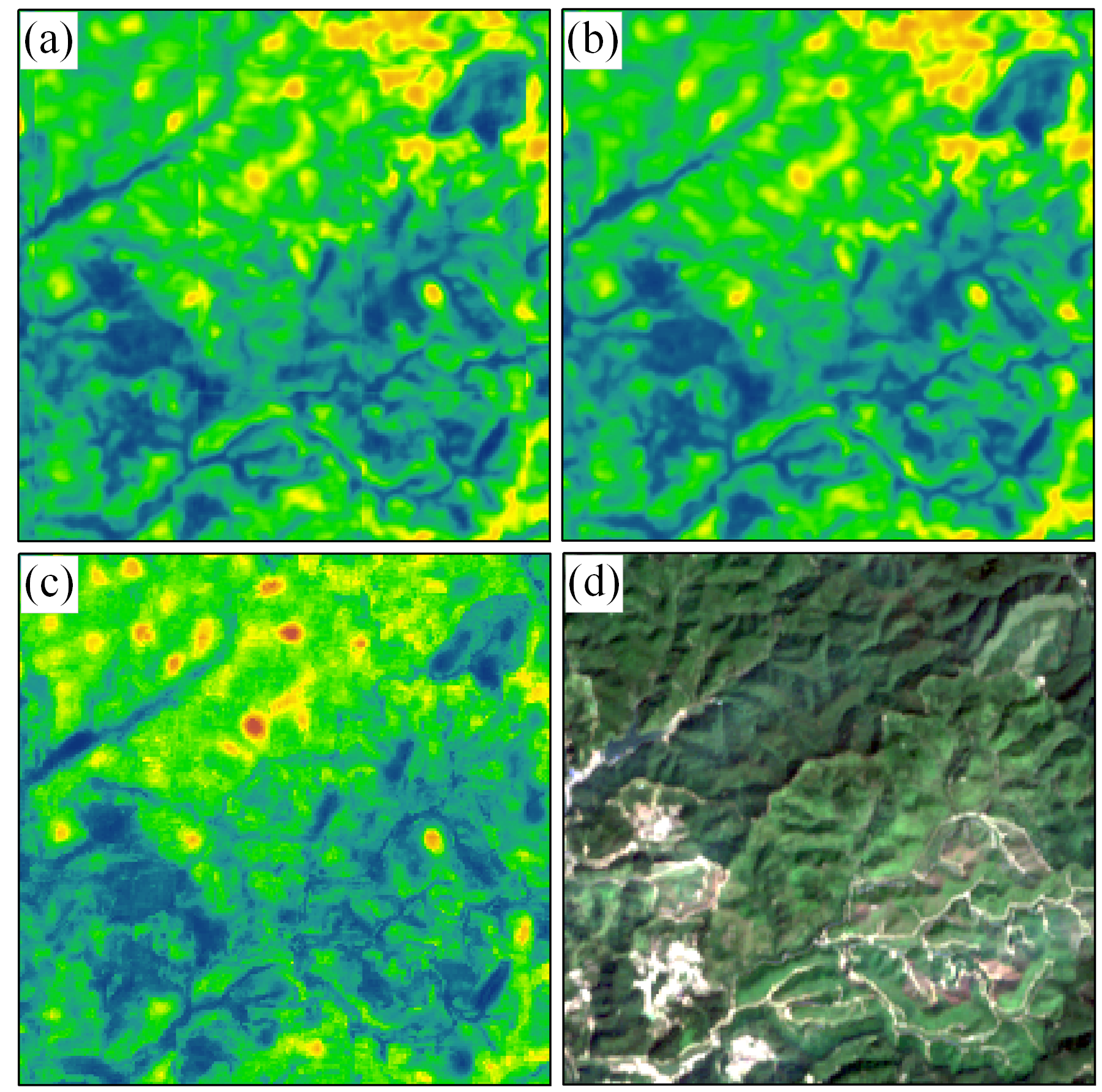}};
        \end{tikzpicture}
    \end{subfigure}%
\caption{Local AGB mapping results for model comparison and boundary effect mitigation demonstration. (a) Localized enlargement of AGB map generated by AU, (b) The refined AGB map by AU after mitigating boundary effects, (c) AGB map produced by RF, (d) Sentinel-2 true-color composite image.}
\label{fig: Boundary effect}
\end{figure}

The AGB map initially generated by AU exhibited boundary effects as shown in (Fig. \ref{fig: Boundary effect} (a)). Discontinuities and inconsistent values can be observed along the patch edges, likely resulting from the lack of contextual information exchange between neighboring image patches during the per-patch based training. To mitigate this deficiency, the boundary effect removal approach laid out in Section \ref{sec: Mitigating boundary effects} of the methodology was applied. As illustrated subsequently in Fig. \ref{fig: Boundary effect} (b), after undergoing the mitigation process, these boundary effects are no longer discernible. This indicates that the overlapping inference and averaging technique helped improve inter-patch coherence by reducing abrupt changes at patch boundaries caused by the boundary effect. The processed results after boundary effect removal clearly delineated the spatial variability and transitions in AGB levels across the landscape. The processed results demonstrated the road network, shown in the lower half of Fig. \ref{fig: Boundary effect} (d), as a consistent low AGB region, accurately capturing its distinct linear footprint. In contrast, the representation by the Random Forest in Fig. \ref{fig: Boundary effect} (c) did not elucidate the feature as distinctly.

\subsubsection{Spatial distribution of AGB and uncertainty map}
\begin{figure}[H]
    \centering
    \begin{subfigure}{1.0\textwidth}
        \centering
        \begin{tikzpicture}
            \node[anchor=south west,inner sep=0] (image) at (0,0) {\includegraphics[width=\textwidth]{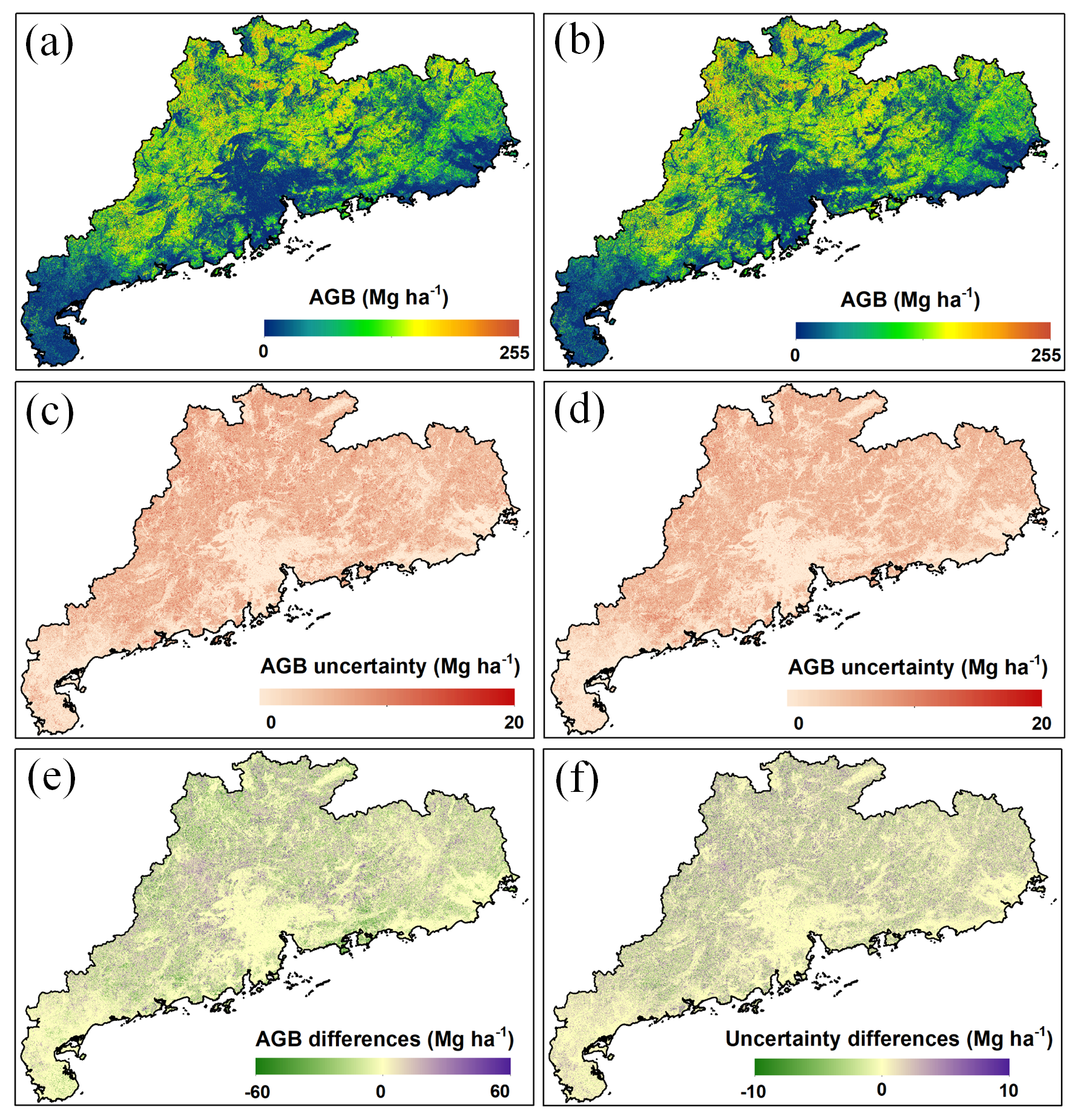}};
        \end{tikzpicture}
    \end{subfigure}%
\caption{Comparative analysis of AGB maps and uncertainty maps at 10m resolution for AU and RF. (a) AU and (b) RF AGB maps, (c) AU uncertainty and (d) RF uncertainty maps, (e)AGB differences, (f) uncertainty differences.}
\label{fig: results distributions}
\end{figure}

The prediction and uncertainty maps by AU and RF at 10 m resolution are shown in Fig. \ref{fig: results distributions}, all the maps have been masked by the ESA forest cover product \citep{zanaga2022esa}. It can be observed that the forest AGB maps from the two models exhibit similar trends, where the northern regions of Guangdong province portray higher AGB, while the southern coastal areas manifest lower AGB values. The uncertainty associated with the forest AGB estimates exhibited a proportional relationship with the predicted AGB values; higher uncertainty was generally observed in areas with greater estimated AGB. The AU model estimated the average forest AGB in Guangdong province to be 102.18 Mg ha\textsuperscript{-1} with an average uncertainty of 5.51 Mg ha\textsuperscript{-1}, while the RF model estimated it to be 104.84 Mg ha\textsuperscript{-1} with an average uncertainty of 5.12 Mg ha\textsuperscript{-1}. At the level of mean values, the estimations from both models appear closely aligned, yet the maps of AGB differences and uncertainty differences reveal distinct discrepancies in AGB and uncertainty between the two models (Fig. \ref{fig: results distributions} (e) (f)).

\begin{figure}[H]
    \centering
    \begin{subfigure}{1.0\textwidth}
        \centering
        \begin{tikzpicture}
            \node[anchor=south west,inner sep=0] (image) at (0,0) {\includegraphics[width=\textwidth]{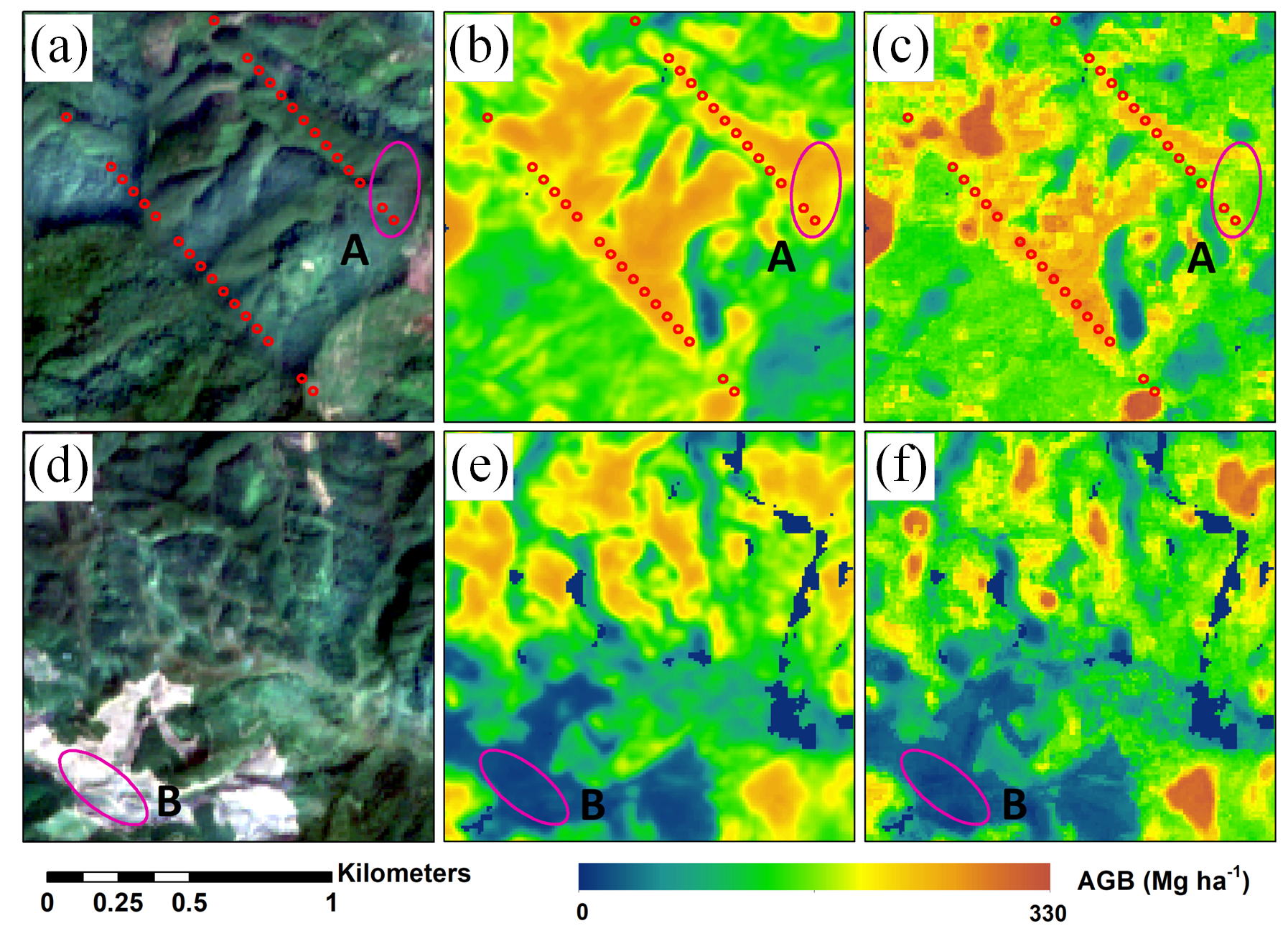}};
        \end{tikzpicture}
    \end{subfigure}%
\caption{Comparison of the AGB maps generated by AU and RF. (a) (d) Sentinel-2 true-color images, (b) (e) AGB maps generated by AU, (c) (f) AGB maps generated by RF. The red circles represent GEDI footprints.
}
\label{fig: Small area}
\end{figure}

To better visualize the the spatial distribution differences between the AGB maps generated by AU and RF, we created enlarged images of specific areas (Fig. \ref{fig: Small area}). Compared to the AGB maps generated by RF (Fig. \ref{fig: Small area} (c) (f)), those produced by AU (Fig. \ref{fig: Small area} (b) (e)) appear to be smoother. The RF-derived maps exhibit fewer high-value areas than those from AU, but they do contain some extremely high values. This is because the multi-scale contextual awareness imparted through the attention mechanism in AU enables modeling of more globally continuous AGB patterns. In contrast, the local splitting criteria used for RF tends to create AGB maps with scattered fragmented segments. Within Area A, the mean AGB derived from two GEDI footprints is 298.50 Mg ha\textsuperscript{-1}. The AU model estimated a mean AGB of 232.59 Mg ha\textsuperscript{-1} for this region, while the RF model estimated a value of 176.77 Mg ha\textsuperscript{-1}. Both models underestimated the AGB for Area A, as is evident in the scatter plot \ref{fig: model performance mean values}. However, the AU predictions are moderately closer to the GEDI reference, outperforming RF in this local region. Area B represents a non-forest region that remains after masking with the ESA forest cover product. Both models successfully characterized these zones by predicting low AGB values, as expected for non-forest land cover types. The AU model estimated a mean AGB of 16.16 Mg ha\textsuperscript{-1} across area B, while the RF model predicted a slightly higher mean AGB of 19.54 Mg ha\textsuperscript{-1}. The lower values from the AU network demonstrate its superior performance in representing low AGB conditions compared to RF.

\subsubsection{Impact of AU architecture depth on results}
To investigate the impact of network depth, we trained and evaluated AU models with 2, 3, 4, and 5 layers using the same experimental setup. The findings indicated marginal performance variations across different depths (Table. \ref{tab:unet_depth}). Specifically, the 4-layer model, which was employed in this study, yielded results closely aligned with its 3-layer and 5-layer counterparts. Our analysis suggests that the dataset used in this study exhibits insensitivity to the depth of the AU architecture. This implies that the intricate spatial patterns and features necessary for accurate biomass estimation may be sufficiently captured even with a reduced or increased number of layers. However, it's crucial to acknowledge that while our dataset demonstrated such characteristics, the insensitivity might not persist across diverse or more complex datasets.

\begin{table}[H]
    \centering
    \caption{Performance metrics for different depths of attention UNet.}
    \label{tab:unet_depth}
    \begin{tabular}{lcc}
        \toprule
        AU Depth & R$^2$ & RMSE (Mg ha\textsuperscript{-1})\\
        \midrule
        2 layers & 0.66 & 43.76 \\
        3 layers & 0.66 & 43.68 \\
        4 layers & 0.66 & 43.66 \\
        5 layers & 0.66 & 43.67 \\
        \bottomrule
    \end{tabular}
\end{table}

\section{Discussion}
\subsection{Contributions of AU}
AU performs best on the test dataset, exhibiting the highest R$^2$ value, alongside the lowest recorded RMSE and bias. For AGB mapping, high R$^2$ and low RMSE are desired to ensure precision, while low bias is required to avoid systematic under- or over-estimation across the entire range of AGB values. However, the use of deep learning methods for forest AGB estimation with GEDI data was not invariably superior to the RF approach. For instance, UNet exhibited a higher bias compared to RF, and ResNet101 showed inferior performance to RF in terms of R$^2$ and RMSE. The underlying reason for this disparity is that UNet lacks the capability to dynamically assign weight to the importance of different feature regions. Concurrently, ResNet is dependent solely on its sequentially stacked convolutional layers and lacks the skip connections inherent in UNet, which are crucial for amalgamating local and global contextual information.

Additionally, we employ a fully connected layer to transform the corresponding remote sensing imagery of each GEDI footprint into a patch suitable for deep learning applications. This methodology inherently precludes the utilization of spatial information within the data. Despite this limitation in exploiting spatial contextual cues inherent in the original data, the AU model on such transformed data still achieved superior accuracy over RF. This suggests that the superior performance of AU can be attributed not only to its effective utilization of spatial information between pixels but also to its exemplary integration of the advantages of attention mechanisms and the UNet architecture. The attention mechanism selectively emphasizes salient features and suppresses irrelevant ones, while the UNet encoder-decoder structure retains local and fine-grained spatial information through the skip connections. This allows AU to effectively learn multi-scale spatial features from the remote sensing inputs. Furthermore, both the Forest AGB map and the uncertainty map demonstrating high consistency with the widely-used RF attest to the reliability of the AU model. In summary, the AU model demonstrated good performance, proving to be a powerful tool for regression tasks, such as forest AGB estimation.

\subsection{Efficiency analysis}
Apart from accuracy, the efficiency of the models is also crucial, especially when deployed over large areas. We compared the training time of all the models, and the time taken to predict AGB across the entire province for the AU and RF. Both of the experiments were conducted on the server mentioned in Section \ref{sec: Model setting}. The time consumption for training of all models, conducted through 5-fold cross-validation, along with the time spent by the AU and RF models in predicting AGB across the entire province using five different trained models, is shown in Table \ref{tab:speed}. During the training phase, it was observed that most deep learning models typically converged before or around 90 epochs; consequently, 120 epochs were selected as the maximum for training to ensure thorough learning without unnecessary computational expense. The deep convolutional networks entailed substantially longer training times compared to the conventional RF algorithm. Among the deep models, AU-FC took the most extensive computational time during training because each GEDI footprint was transformed into an individual patch. Our primary model, AU, utilized the second-longest duration for training, and notably, in comparison to UNet, the attention mechanism did not demand significantly additional time. When applying the 5 trained models to predict AGB across Guangdong province, the inference time of the AU and RF models was comparable, even though parallel processing was utilized for the RF implementation. This is because the AU network utilized efficient GPU hardware acceleration, while the RF model leveraged multi-core CPU parallelism by distributing predictions across threads. Overall, the total training and inference time of AU was not substantially greater than the RF model. With future upgrades in hardware acceleration, such as utilization of high-performance computing clusters, the computational expenses of AU can potentially be further reduced. In particular, the inference speeds already matched those of RF, demonstrating the viability of deploying well-trained deep networks for expansive spatial and temporal predictions.

\begin{table}[H]
\centering
\caption{Time consumption of the models (minutes)} 
\begin{tabular}{ccc}
\toprule
Model & Training time & Prediction time \\ \hline 
AU& 587 & 1318 \\ 
RF& 201 & 1379 \\
AU-FC& 1570 & - \\
UNet& 560 & - \\
ResNet101& 350 & - \\
\bottomrule 
\end{tabular}
\label{tab:speed}
\end{table}

\subsection{Potential of deep learning for forest AGB estimation}
The confluence of unprecedented data sources, augmented computational capabilities, and the latest advancements in deep learning presents exhilarating new opportunities to enhance our understanding of forest biomass derived from data. We have demonstrated through the use of FC layers that utilizing spatial information can improve the accuracy of AGB estimation. One of the advantages of deep learning is its capability to effectively utilize spatial information, presenting a promising future in the quantitative characterization of forest carbon dynamics. While advanced deep learning algorithms are continuously emerging in the realm of deep learning \citep{he2022masked, liu2021swin}, they necessitate further refinement and adaptation for the field of remote sensing. This is particularly true for regression analysis tasks, such as estimating forest biomass using remote sensing data.

\section{Conclusions}
The deep learning approach we proposed offers an innovative computational methodology for estimating forest AGB, leveraging GEDI and remote sensing imagery. This study demonstrates the feasibility of mapping forest aboveground biomass (AGB) over large areas by leveraging publicly accessible remote sensing data sources, including Sentinel-1, Sentinel-2, and ALOS-2 PALSAR-2, in conjunction with advanced deep learning methodologies and the open GEDI dataset as a reference. Our model significantly reduces the overall bias in the estimation of forest AGB, concurrently enhancing R$^2$ and reducing RMSE. Moreover, it exhibits an enhanced capability to discern regions characterized by high and low AGB values with heightened precision. Given sufficient training data, such deep learning techniques may better capture complex ecological gradients and provide enhanced generalization across different forest types and disturbance conditions. Given the importance of AGB as an indicator of climate change impacts and mitigation, these methods for continuous mapping from freely available data represent a valuable tool for understanding and addressing pressing environmental challenges.

However, it is crucial to note that not all deep learning approaches can guarantee an enhancement in accuracy. Further research is imperative to devise suitable network architectures, optimize training strategies, including patch size, and to validate the methodologies across varied forest environments. Although deep learning algorithms typically demand more computational resources, it's worth noting that AGB maps do not necessarily require a high temporal resolution. In most cases, annual AGB maps are sufficient. Overall, the integration of emerging Lidar datasets like GEDI with optical and SAR data through advanced deep learning has significant potential for improving large scale forest AGB estimation. 

\section{Code availability}
The code to train deep learning models and predict AGB maps will be available online at https://github.com

\section{Acknowledgements}
C.M.R. was supported by the NERC funded SECO project: NE/T01279X/1. The field work was supported by Davis Expedition Fund, Elizabeth Sinclair Irvine Bequest and Centenary Agroforestry 89 Fund, Moray Endowment Fund and Meiklejohn fund. 

\bibliographystyle{elsarticle-harv} 
\bibliography{references}





\end{document}